%% file: main.tex
\documentclass[runningheads]{llncs}

\usepackage{eccv}

\usepackage{eccvabbrv}

\usepackage{graphicx}
\usepackage{wrapfig}
\usepackage{wrapfig}
\usepackage{booktabs}
\usepackage{multirow}
\usepackage{pgfplots}
\usepackage{algorithm}
\usepackage{algorithmic}
\usepackage{amsmath}
\usepackage{amsfonts}
\pgfplotsset{compat=1.18}

\usepackage[accsupp]{axessibility}  % Improves PDF readability for those with disabilities.

\usepackage{hyperref}

\usepackage{orcidlink}

\begin{document}

% ---------------------------------------------------------------
% TODO REVIEW: Replace with your title
\title{Anchored Video Generation: Decoupling Scene Construction and Temporal Synthesis in Text-to-Video Diffusion Models} 

% TODO REVIEW: If the paper title is too long for the running head, you can set
% an abbreviated paper title here. If not, comment out.
\titlerunning{Anchored Video Generation}

% TODO FINAL: Replace with your author list. 
% Include the authors' OCRID for the camera-ready version, if at all possible.
% \author{First Author\inst{1}\orcidlink{0000-1111-2222-3333} \and
% Second Author\inst{2,3}\orcidlink{1111-2222-3333-4444} \and
% Third Author\inst{3}\orcidlink{2222--3333-4444-5555}}
\author{Mariam Hassan$^*$, Bastien Van Delft$^*$, Wuyang Li, Alexandre Alahi
% {\tt\small firstauthor@i1.org}
% For a paper whose authors are all at the same institution,
% omit the following lines up until the closing ``}''.
% Additional authors and addresses can be added with ``\and'',
% just like the second author.
% To save space, use either the email address or home page, not both
% \and
% Second Author\\
% Institution2\\
% First line of institution2 address\\
% {\tt\small secondauthor@i2.org}
}

% TODO FINAL: Replace with an abbreviated list of authors.
\authorrunning{M. Hassan et al.}
% First names are abbreviated in the running head.
% If there are more than two authors, 'et al.' is used.

% TODO FINAL: Replace with your institution list.
\institute{École Polytechnique Fédérale de Lausanne (EPFL)  \\
\email{firstname.lastname@epfl.ch}}

\maketitle

\renewcommand\thefootnote{}
\footnotetext{$^*$ Equal Contribution}

\input{sec/0_abstract}    
\input{sec/1_intro}
\input{sec/2_related_works}
\input{sec/3_methods}
\input{sec/4_experiments}

\input{sec/5_discussion}
\section{Acknowledgments}
This work was supported as part of the Swiss AI Initiative by a grant from the Swiss National Supercomputing Centre (CSCS) under project ID a144. The research was supported by Innosuisse – the Swiss Innovation Agency (Ref. 127.265 IP-ICT).

% \clearpage  % TODO FINAL: This \clearpage needs to be removed from both review and camera-ready versions.

% \section*{Acknowledgements}
% Please insert your acknowledgments here.

% ---- Bibliography ----
%
% BibTeX users should specify bibliography style 'splncs04'.
% References will then be sorted and formatted in the correct style.
%
\bibliographystyle{splncs04}
\bibliography{main}
\input{sec/X_suppl}
\end{document}

%% file: sec/0_abstract.tex
\begin{abstract}
State-of-the-art Text-to-Video (T2V) diffusion models can generate visually impressive results, yet they still frequently fail to compose complex scenes or follow logical temporal instructions. In this paper, we argue that many errors, including apparent motion failures, originate from the model's inability to construct a semantically correct or logically consistent initial frame.
We introduce Anchored Video Generation (AVG), a modular pipeline that decouples these tasks by decomposing the Text-to-Video generation into three specialized stages: (1) Reasoning, where a Large Language Model (LLM) rewrites the video prompt to describe only the initial scene, resolving temporal ambiguities; (2) Composition, where a Text-to-Image (T2I) model synthesizes a high-quality, compositionally-correct anchor frame from this new prompt; and (3) Temporal Synthesis, where a video model, finetuned to understand this anchor, focuses its entire capacity on animating the scene and following the prompt. Our approach sets a new state-of-the-art on the T2V CompBench benchmark and significantly improves all tested models on VBench2. Furthermore, we show that visual anchoring allows us to cut the number of sampling steps by 70\% without any loss in performance. AVG offers a simple yet practical path toward more efficient, robust, and controllable video synthesis. Project website: \url{https://vita-epfl.github.io/AVG/}

% Text-to-video (T2V) diffusion models have recently achieved strong progress in visual fidelity, motion continuity, and prompt controllability, yet they remain fragile: even simple prompts often fail due to incorrect or logically inconsistent first-frame construction. Our analysis shows that these failures stem primarily from conflating two distinct tasks—static scene composition and temporal synthesis—within a single diffusion process.
% We introduce Factorized Video Generation (FVG), a modular pipeline that explicitly separates T2V generation into three stages: (1) reasoning, where a language model rewrites the prompt into its correct initial state; (2) composition, where a text-to-image model produces a grounded anchor frame; and (3) animation, where a video diffusion model generates motion conditioned on the anchor and the original prompt.
% Across multiple open-source T2V models, FVG yields substantial gains on T2V-CompBench and VBench2.0 and establishes a strong new baseline for grounded video generation. FVG also enables significant reductions in sampling steps without degrading quality. These results show that explicit scene grounding, rather than larger end-to-end models, is key to more reliable T2V diffusion.
\end{abstract}

%% file: sec/1_intro.tex
\section{Introduction}
\label{sec:intro}
\begin{figure}[t]
\label{fig:pull}
\centering
\includegraphics[width=\linewidth]{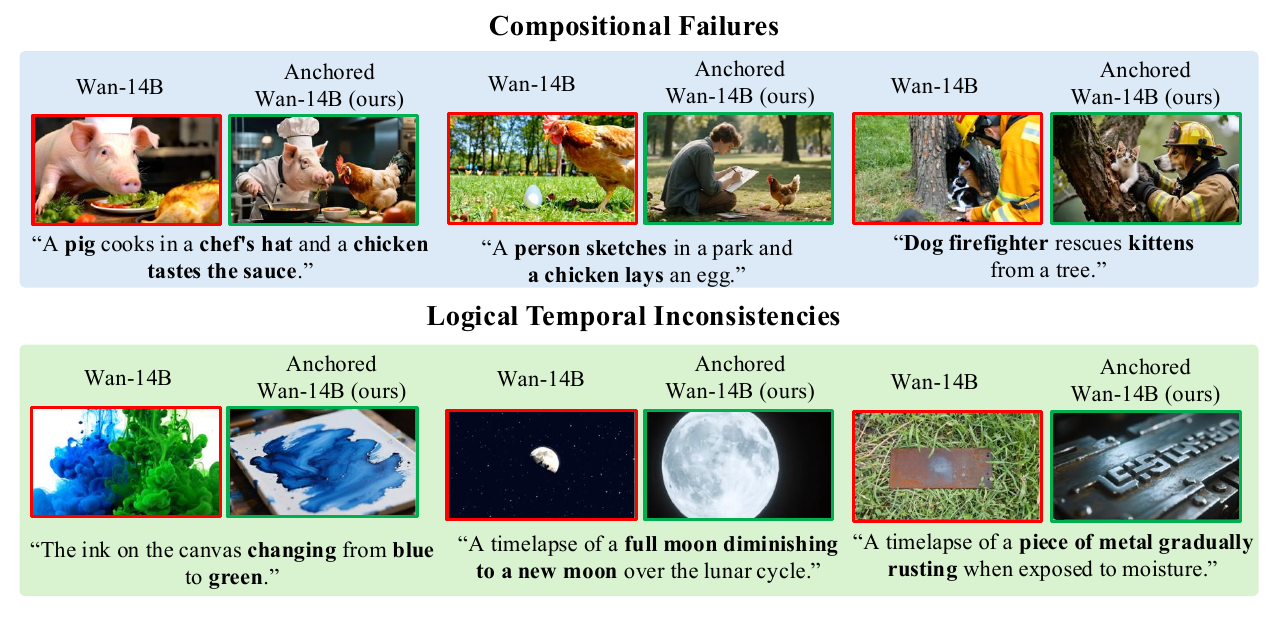}
\caption{\textbf{Example failure modes in SoTA video generative models.} We compare the first frame of videos from T2V Wan 2.2~\cite{wan2025wan} and our \textbf{Anchored} Wan 2.2 under the AVG pipeline. T2V Wan 2.2 composes scenes incorrectly and exhibits logical temporal inconsistencies, struggling to establish coherent scene structure without explicit visual grounding.}
\label{fig:pull_figure}
\end{figure}

Recent advances in text-to-video (T2V) diffusion models have substantially improved visual fidelity, temporal consistency, and prompt adherence~\cite{bar2024lumiere, wan2025wan, yang2025cogvideox,liu2024sora,videoworldsimulators2024,blattmann2023stable,ho2022video,lin2024open}. However, despite these achievements, current systems exhibit notable brittleness: they often generate visually convincing videos that nevertheless violate basic semantic constraints in the prompt, particularly when the prompt requires binding specific actions to specific entities or maintaining a coherent logical sequence of events. These failures highlight persistent limitations in how existing models \textbf{jointly reason about language, scene structure, and temporal evolution} in a unified video generator.

Prior works have documented these compositional failures \cite{tian2024videotetris,sun2025t2v,xu2025comocompositionalmotioncustomization,huang2024vbench, zheng2025vbench20advancingvideogenerationbenchmark}, showing that T2V models frequently misinterpret or only partially follow user instructions, especially when the prompts deviate from the curated, syntactically clean data used during training. A common mitigation is the \textbf{prompt extension}, which expands short or underspecified prompts using Large Language Models (LLMs)~\cite{cheng2025vpo, lee2024videorepair, huang2024storydiffusion}. While such rewriting can increase textual specificity, it does not address the underlying representational issue: the model may still construct an incorrect scene or misrepresent the entities and relations described, regardless of how detailed the rewritten prompt becomes.

Our analysis across multiple state-of-the-art systems reveals that many errors originate in the model’s representation of the scene itself. In \Cref{fig:pull_figure}, we observe that existing models routinely \textbf{(1) miscompose visual elements of the scene}, and \textbf{(2) generate initial states incompatible with the requested temporal evolution}. Once the scene composition violates the prompt semantics, the temporal modeling fails. For example, if the model wrongly understands the context by assuming a “chicken" as a “roasted chicken" (see 1st sample), it becomes impossible to satisfy the prompt “a chicken tasting the sauce,” regardless of the strength of the motion model. 
This insight indicates a critical gap:
Current T2V models may have \textbf{over-entangled} the two distinct tasks, static scene construction and temporal synthesis, yet are not structurally equipped to handle the first reliably.

To delve into and understand this gap, we explore a diagnostic study by comparing Text-to-Video (T2V), Image-to-Video without text prompts (I2V), and Image-to-Video with text prompts (I2V+text) models in \Cref{tbl:fid-fvd}, revealing two observations. T2V models frequently produce distributionally shifted frames yet still achieve FVD scores comparable to I2V models, indicating that \textbf{their motion modeling remains reasonably natural even when scene fidelity is relatively poor}. I2V models exhibit the complementary behavior, strong FID from accurate initial scenes and weaker temporal coherence, while I2V+text balances both aspects. This contrast suggests a \textbf{structural mismatch in current T2V models}: scene grounding and temporal synthesis are two distinct modeling challenges, yet existing architectures attempt to learn both simultaneously within a single model.

These observations motivate us to revisit T2V. \emph{Rather than formulating it as a single task, we decompose it into two fundamentally sub-problems: specifying the initial scene and generating its temporal evolution.} A natural way to operationalize this separation is through a visual anchor defining the intended starting state of the video. However, such an anchor cannot be generated by simply running a Text-to-Image (T2I) model on the original video prompt, which often contains multi-stage actions or temporal transformations. Obtaining an appropriate anchor requires first clarifying the prompt’s intended initial condition.

Based on this insight, we propose Anchored Video Generation (AVG), a pipeline that explicitly decomposes T2V generation into specialized stages designed to address each subproblem independently:
\textbf{(1) Reasoning:} A Large Language Model (LLM) rewrites the video prompt into an explicit description of the initial scene, removing temporal transitions and eliminating logical ambiguities.
\textbf{(2) Composition:} A T2I model uses this refined prompt to synthesize a high-quality, semantically grounded anchor frame.
\textbf{(3) Temporal synthesis}: A video diffusion model conditions on both the anchor and the original prompt, focusing exclusively on generating coherent temporal dynamics.
We refer to this instantiated model variant as \textbf{anchored T2V}, as it augments a standard T2V backbone with an explicit intermediate visual anchor that establishes the intended scene composition.
The contributions of this work are as follows:

\begin{itemize}
\item Through qualitative and quantitative evaluation, we show that scene composition is one of the primary failure modes in existing T2V systems.
\item We introduce Anchored Video Generation, a simple and modular three-stage pipeline that separates reasoning, scene composition, and temporal synthesis, enabling improved scene grounding and temporal coherence.
\item We demonstrate that this decomposed approach sets a new state-of-the-art on the CompBench benchmark \cite{sun2025t2v} and significantly improves all tested models on VBench2~\cite{zheng2025vbench20advancingvideogenerationbenchmark}.
\item We show that visual anchoring enables significant reductions in sampling steps without degrading output quality, providing a practical path toward more efficient video generation.
\end{itemize}

\noindent We will also release our models, code, benchmark prompts and image anchors generations to support reproducibility and facilitate further research.

%% file: sec/2_related_works.tex
\begin{table}[t]
\centering
\caption{Diagnostic study in understanding the over-entangled gap. FID/FVD of WAN 2.2 (5B) with 5k samples on two video generation datasets. We compare T2V, I2V, and I2V+text generations. I2V+text consistently achieves the best FID/FVD.}
\label{tbl:fid-fvd}

\setlength{\tabcolsep}{4pt}

\begin{tabular}{llrrrrrr}
\toprule
& & \multicolumn{3}{c}{\textbf{VidGen}~\cite{wang2024vidgen}} 
  & \multicolumn{3}{c}{\textbf{OpenVid-1M}~\cite{nanopenvid}} \\
\cmidrule(lr){3-5} \cmidrule(lr){6-8}
& \textbf{Metric} 
  & \textbf{T2V} & \textbf{I2V} & \textbf{I2V+Text} 
  & \textbf{T2V} & \textbf{I2V} & \textbf{I2V+Text} \\
\midrule
& FID & 14.10 & 10.12 & \textbf{5.25}
      & 17.28 & 8.56 & \textbf{3.33} \\
& FVD & 178.76 & 142.12 & \textbf{34.78}
      & 28.99 & 124.58 & \textbf{27.08} \\
\bottomrule
\end{tabular}

\end{table}

\section{Related Work}
\label{sec:related}

\noindent\textbf{Image Generative Models.} With systematic scaling of architectures and datasets, state-of-the-art image generation models~\cite{wu2025qwen,rombach2022high,podell2023sdxl,li2024hunyuan,esser2024scaling,chen2023pixart,wang2024emu3}, exemplified by the 20B-parameter Qwen-Image~\cite{wu2025qwen}, achieve photorealistic synthesis with fine-grained controllability that flexibly adapts to user specifications. These models exhibit robust text-image alignment, semantic coherence, and comprehensive multimodal editing capabilities~\cite{brooks2023instructpix2pix,huang2025diffusion} in an efficient manner.

\noindent\textbf{Video Generative Models.} In parallel, video models~\cite{liu2024sora,videoworldsimulators2024,yang2025cogvideox,blattmann2023stable,ho2022video,lin2024open} have also made great progress in generating temporally coherent short contents, which can also be extended to minute-length long videos~\cite{zhang2025packing,li2025stable,lu2024freelong,team2025longcat,wang2024loong,ge2022long}. However, compared with the image counterpart, introducing the extra temporal dimension presents fundamental challenges: Video data has \emph{{exponentially higher dimensionality}} thus requires substantially richer text-video pairs, also leading to computational restrictions on model capacity. For example, current video models, such as Wan~\cite{wang2025wan} (14B), remain significantly smaller than image counterparts like Qwen-Image~\cite{wu2025qwen} (20B), limiting the semantic understanding. Consequently, existing models exhibit suboptimal text-following capabilities~\cite{huang2023vbench,liu2024redefining}, often resulting in missing concepts, temporal inconsistencies, and reduced prompt adherence compared to the image-generation counterparts.

% \noindent\textbf{Spatial-Temporal Factorization.} To address the complexity in video-centric learning, factorization~\cite{tran2018closer} has become a fundamental architectural strategy~\cite{liu2022video,bertasius2021space,lin2019tsm,ma2024latte}, which decouples spatial and temporal components to enable efficient video representation. Some works~\cite{chen2023videocrafter1,videoworldsimulators2024} extend this paradigm via hierarchical factorization across multiple scales. At a higher level, Emu Video~\cite{girdhar2024factorizing} similarly explores factorizing text-to-video generation through an explicit intermediate image. While Emu Video studies this idea through a dedicated factorized generator, our method is designed as a lightweight retrofit for existing pretrained T2V backbones, making the approach more modular and easier to integrate in practice. Existing architectural factorization approaches can also compromise spatial-temporal coupling under limited model capacity, leading to suboptimal results when data is scaled~\cite{ma2024latte}. \emph{Different from these lines of work, we lift factorization to a higher, task-oriented level}, thereby preserving model capacity while achieving stronger text-following capabilities in video generation.

\noindent\textbf{Spatial-Temporal Factorization.} To address the complexity in video-centric learning, factorization~\cite{tran2018closer} has become a fundamental architectural strategy~\cite{liu2022video,bertasius2021space,lin2019tsm,ma2024latte}, which decouples spatial and temporal components to enable efficient video representation. Some works~\cite{chen2023videocrafter1,videoworldsimulators2024} extend this paradigm via hierarchical factorization across multiple scales. At a higher level, Emu Video~\cite{girdhar2024factorizing} similarly explores factorizing text-to-video generation through full training with an explicit intermediate image. While Emu Video studies this idea through a dedicated factorized generator, our method is designed as a lightweight retrofit for existing pretrained T2V backbones, making the approach more modular and easier to integrate in practice. Existing architectural factorization approaches can also compromise spatial-temporal coupling under limited model capacity, leading to suboptimal results when data is scaled~\cite{ma2024latte}. \emph{Different from these lines of work, we lift factorization to a higher, task-oriented level}, thereby preserving model capacity while achieving stronger text-following capabilities in video generation.
\vspace{-5mm}

%% file: sec/3_methods.tex
\begin{figure*}[t]
\label{fig:method}
\centering
\includegraphics[width=\textwidth]{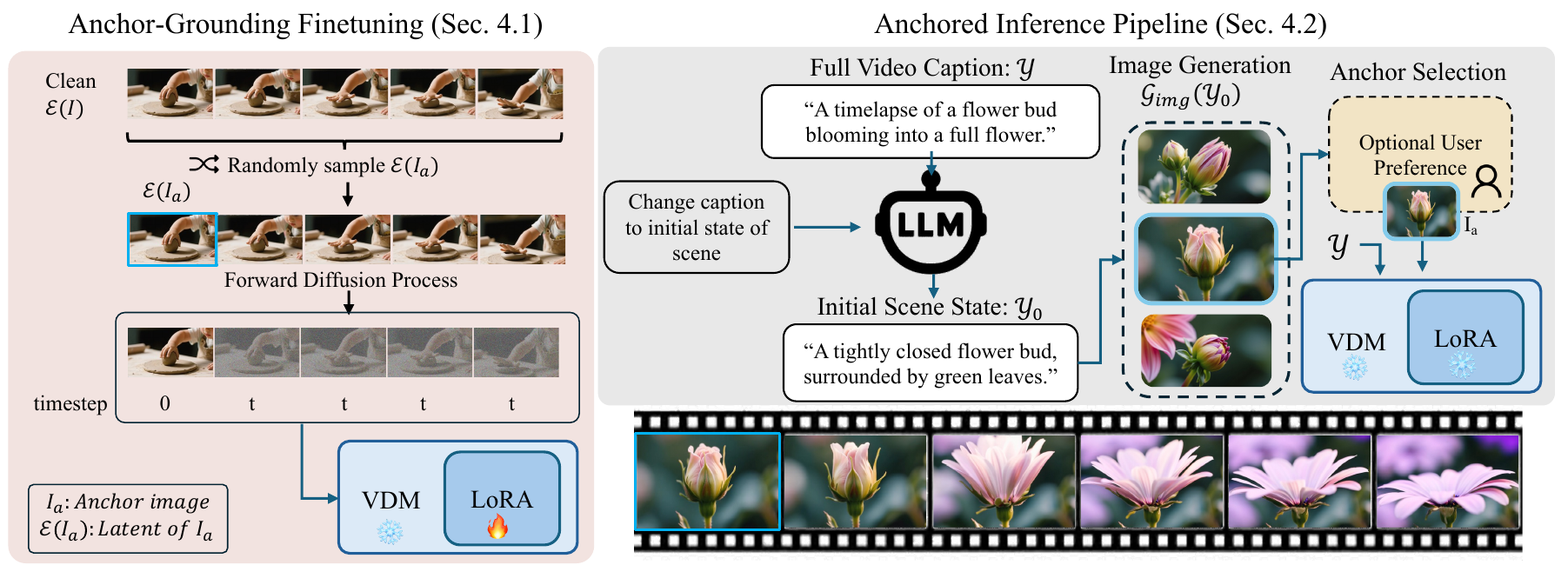}
\caption{Overview of the finetuning and inference pipelines in AVG. Anchor-Grounding Finetuning: the Video Diffusion Model (VDM) is trained to follow a visual anchor by injecting a clean image latent at a randomly chosen frame position and setting its diffusion timestep to $t=0$. Lightweight LoRA finetuning teaches the model to treat this clean frame as a fixed scene constraint that guides the rest of the video. Anchored Generation Pipeline: an LLM rewrites the video prompt into a first-frame scene description to generate an anchor image. The anchor is then injected into the VDM with its timestep fixed to $0$.}
\label{fig:overview}
\end{figure*}

\section{Preliminaries}
\label{sec:preliminaries}

\paragraph{Flow Matching for Denoising.}
Recent video diffusion models are grounded in the flow-matching formulation~\cite{lipman2022flow,liu2023flowmatching,kim2024flowvid}, which views denoising as learning a continuous-time flow field that maps a simple prior distribution, typically Gaussian, to the target data distribution.
Given video latents $\mathbf{z}_0 \sim \mathcal{N}(0, \mathbf{I})$ and data samples $\mathbf{z}_1 \sim p_{\text{data}}(\mathbf{z})$, the model learns a vector field $\mathbf{v}_\psi(\mathbf{z}_\tau, \tau)$ that transports intermediate states $\mathbf{z}_\tau = \tau \mathbf{z}_1 + (1-\tau)\mathbf{z}_0$ along time $\tau \in [0,1]$:
\begin{equation}
\mathcal{L}(\psi) = \mathbb{E}_{\tau, \mathbf{z}_0, \mathbf{z}_1}
\big\|
\mathbf{v}_\psi(\mathbf{z}_\tau, \tau) - (\mathbf{z}_1 - \mathbf{z}_0)
\big\|_2^2.
\label{eq:flow}
\end{equation}
In practice, this formulation generalizes the discrete-step denoising process of diffusion models, where the continuous flow is approximated by $T$ discrete denoising steps. A higher number of steps yields finer approximations of the flow trajectory but at a higher computational cost. Our study later explores how visual grounding improves robustness even when $T$ is significantly reduced.

\paragraph{Conditioning Mechanisms.}
Conditioning guides the generation toward desired semantics or structure. Text prompts are typically encoded into embeddings $h_y$ via a transformer encoder (e.g., T5~\cite{raffel2020t5}) and injected into the model through cross-attention:
\begin{equation}
\mathbf{h}' = \mathrm{Attention}(\mathbf{Q}=\phi(\mathbf{z}_t),\,
\mathbf{K}=h_y,\,\mathbf{V}=h_y),
\end{equation}
where $\phi(\mathbf{z}_t)$ denotes the query features from the current latent.
For image conditioning, a reference frame $\mathbf{x}_0$ or its encoded features $h_{\mathbf{x}_0}$ are concatenated with the video latent representation, either channel-wise or in the temporal dimension:
\begin{equation}
\mathbf{z}'_t = [\,\mathbf{z}_t ; h_{\mathbf{x}_0}\,],
\end{equation}
allowing the model to anchor spatial structure before learning temporal evolution.
This combination—text via cross-attention and image via concatenation is referred to as image-to-video (I2V) diffusion with text conditioning, enabling the model to generate semantics and scene layout jointly. However, such joint conditioning typically requires extensive retraining and substantial computational resources to align multimodal representations effectively.

\section{Methods}
\label{sec:method}
The goal of this work is to investigate how grounding text-to-video (T2V) diffusion models with at least one visual anchor affects their performance on complex compositional tasks. By introducing a visual anchor, we aim to decouple video generation into two subtasks, scene context construction and temporal motion modeling, and evaluate whether this separation improves the model’s understanding of scenes, objects, and their interactions. Our focus extends beyond visual quality to assess reasoning-oriented capabilities of video diffusion models. To this end, we compare standard T2V models with their text–image–to–video (I2V+text) counterparts across benchmarks designed to evaluate compositional video generation. Furthermore, to enable a fair comparison within the same architecture, we propose dedicated finetuning and inference pipelines that integrate a visual anchor into a pure T2V model. This allows the same backbone to operate either in a standard text-only mode or in an anchor-grounded mode. The following section details these pipelines which are summarized in~\cref{fig:method}.

\subsection{Anchor-Grounding Finetuning}
\label{sec:t2v_to_i2v}

A core capability required by the our approach is the ability to inject a visual anchor into a pretrained T2V diffusion model. We enable this through anchor-grounding finetuning, which teaches the model to treat the provided image as the initial frame of the generated video.

\subsubsection{Latent Injection for Anchor Grounding}
To enforce strong grounding, we directly intervene in the diffusion process by injecting the clean latent of the anchor image into the model’s noisy latent sequence, inspired by diffusion-based inpainting methods~\cite{ti2vzero}.
Specifically, given a noisy video latent $x_t$ containing $T$ frames sampled at the same diffusion timestep $t$, we replace one frame with the clean latent of the anchor image (i.e., at timestep $t=0$), forming a hybrid latent: $T-1$ noisy frames at timestep $t$, 1 clean anchor frame at timestep $0$.

\noindent This hybrid input is fundamentally out-of-distribution for a pretrained T2V model, which is trained exclusively on uniform-timestep sequences. Without any finetuning the method failed as the model failed to adapt to this hybrid input. 
However, we find that the model can reliably learn this behavior with lightweight LoRA finetuning, enabling it to interpret the $t=0$ frame as a fixed visual constraint.

\subsubsection{Training Procedure}
The training procedure for adapting a T2V model to a visually anchored model is detailed in Algorithm \ref{alg:adapt}. The core idea is to create a hybrid input $(z_t, t_{vec})$ that pairs $T-1$ noisy frames at timestep $t$ with one clean anchor frame at $t=0$. The model $\epsilon_\theta$ is then finetuned via LoRA to denoise this out-of-distribution input, learning to treat the $t=0$ frame as a ground-truth anchor. The model remains text-conditioned through the original prompt $y$, while the visual anchor is introduced directly through the latent sequence.

\begin{algorithm}[t]
\caption{Anchor-Grounding Finetuning}
\label{alg:adapt}
\begin{algorithmic}[1]
\REQUIRE Pretrained model $\epsilon_\theta$, Video $z_0$, Text $y$
\\
\STATE $k \sim \text{Uniform}(1, T)$ \COMMENT{Random anchor frame index}
\STATE $z_a \gets z_0[k]$
\STATE $t \sim \text{Uniform}(0, 1)$ 
\STATE $\epsilon \sim N(0, I)$ \COMMENT{Sample noise}
\STATE Construct $z_t$ and $t_{vec}$
\FOR{$j = 1$ to $T$}
    \IF{$j == k$}
        \STATE $z_t[j] \gets z_a$ \COMMENT{Inject clean anchor}
        \STATE $t_{vec}[j] \gets 0$
    \ELSE
        \STATE $z_t[j] \gets (1 - t) \times z_0[j] + t \times \epsilon[j]$
        \STATE $t_{vec}[j] \gets t$
    \ENDIF
\ENDFOR
\STATE $\text{loss} \gets \| \epsilon - \epsilon_\theta(z_t, t_{vec}, y) \|_2^2$
\end{algorithmic}
\end{algorithm}

\subsection{Anchored Generation Pipeline}
At inference time, we first construct a visual anchor from the input text prompt before video sampling. 
Given a video text prompt $y$, we use a large language model (LLM) to generate a plausible detailed description of the scene’s initial state $y_0$, which emphasizes spatial layout, appearance, and object configuration. This reformulated prompt $y_0$ is then passed to a pretrained image generation model $\mathcal{G}_{\text{img}}$ to synthesize the anchor image $I_a = \mathcal{G}_{\text{img}}(y_0)$. 

Optionally, user control can be introduced at this stage to select the preferred visual style or appearance for the anchor. The selected generated image $I_a$ is subsequently encoded into its latent representation $\mathcal{E}(I_a)$. 
Then, we apply this learned behavior to perform anchored generation. We set the anchor frame $k=0$ (first frame) and provide the desired anchor latent $z_a = \mathcal{E}(I_a)$. Then during the denoising process, at each step $t_i$, we reconstruct a hybrid input for the model by replacing the first frame of the current latent $z_{t_i}$ with the clean anchor $z_a$. We similarly construct a hybrid timestep vector $t_{vec}$ where $t_{vec}[0] = 0$ and all other elements are set to $t_i$. The model then predicts the flow using this modified input.

Formally, at each sampling step $t$, the model updates the video latents according to:
\begin{equation}
z_{t_{i-1}} = z_{t_i} - \Delta t \, \mathbf{v}_\theta(z_t, t_{vec}, y),
\end{equation}
where $\mathbf{v}_\theta$ denotes the learned velocity field of the video diffusion model. This ensures that the denoising trajectory evolves consistently from the anchored visual state. 

This inference procedure enables anchor-grounded generation without requiring architectural changes or additional conditioning modules, allowing the model to produce visually consistent and semantically aligned videos while preserving controllability over scene initialization.

\subsection{Implementation Details}

We finetune Wan2.2-14B~\cite{wan2025wan}, CogVideo1.5-5B, and Wan2.1-1B~\cite{wan2025wan} using LoRA modules with a rank of 256, applied to all layers of the diffusion backbone. Finetuning is performed on 5000 randomly sampled videos from the UltraVideo~\cite{xue2025ultravideo} dataset for 6k steps with effective batch size of 16. Training requires around 48 GPU hours for Wan-1B and CogVideo-5B, and 96 GPU hours for Wan-14B. Wan 5B model natively supports both text-only and text–image conditioning; therefore, no finetuning is applied.

%% file: sec/4_experiments.tex
\section{Experiments} \label{sec:exp}
In this section, we start by defining our experimental setup followed by an explanation of the experiments conducted. 

\subsection{Experimental Setup}

% \textbf{Models.} We evaluate four video diffusion models of varying scales and architectures: \textsc{CogVideo 1.5-5B}, \textsc{WAN 2.1-1B}, \textsc{WAN 2.2-5B}, and \textsc{WAN 2.2-14B}. For each model and benchmark category, we generate five random seeds and report the mean score. In the \textsc{TI2V}, the best anchor out of five images across five seeds, to assess sensitivity to the visual conditioning. For the text-only \textsc{T2V} setting, we report results both with and without prompt upsampling. 
To obtain image anchors, each caption is first refined using Qwen2.5-7B-Instruct \cite{qwen2.5vl32binstruct2025} to produce a detailed ``first-image caption'' describing the initial scene state. This refined description is then passed to \textsc{QwenImage}~\cite{qwenimage2025} to synthesize a high-quality anchor image that serves as the first frame for the video diffusion model.

\paragraph{Evaluation Benchmarks.}
We evaluate our models on two text-to-video generation benchmarks.  
T2V-CompBench~\cite{sun2025t2v} measures compositional control over objects, attributes, and actions, assessing how well a model integrates multiple concepts within a coherent scene.  
VBench 2.0~\cite{zheng2025vbench20advancingvideogenerationbenchmark} provides a holistic evaluation across 18 metrics, grouped into five high-level categories: creativity, commonsense reasoning, controllability, human fidelity, and physics. These categories collectively capture both semantic and physical consistency in video generation. 
% Since Vbench requires a lot more to get the results, we only benchmark 3 models on it: Wan5B, Wan14B and Cogvideo 5B.

% We evaluate four models, CogVideo 1.5-5B, WAN 2.1-1B, WAN 2.2-5B, and WAN2.2 14B, on two established video generation benchmarks: T2V-CompBench~\cite{sun2025t2v}, which measures compositional control over objects, attributes, and actions, and VBench 2.0~\cite{zheng2025vbench20advancingvideogenerationbenchmark}, which assesses intrinsic faithfulness across physics, semantics, and prompt adherence. For each method and category, we generate five random seeds and report the mean score. In the TI2V setting, we test five distinct starting images and additionally the best-performing image with five seeds to measure sensitivity to the visual anchor. For T2V, we report results both with and without prompt upsampling. To obtain the image anchor, we refine each caption using Qwen2.5-VL-32B-Instruct~\cite{qwen2.5vl32binstruct2025} to extract a “first-image caption,” which we then pass to QwenImage~\cite{qwenimage2025} to produce a high-quality initial frame.

\subsection{Performance on Complex Tasks}
\textbf{Results on T2V-CompBench:} As shown in~\cref{tbl:compbench} across all models, adding an anchor image consistently improves compositional performances. All smaller anchored models (CogVideo 5B, Wan 5B and Wan 1B)  outperform the larger Wan 14B T2V model. \textbf{Our anchored Wan 5B also outperforms commercial PixVerse-V3 baseline which is the best reported model on the benchmark}. This demonstrates that visual grounding substantially enhances scene and action understanding even in smaller-capacity models. Within each model family, the anchored version outperforms the original model. Notably, our lightweight anchor-grounded LoRA on WAN 14B reaches performance comparable its pretrained I2V 14B variant (0.661 vs. 0.666), despite requiring no full retraining.  \\
% Interestingly, we observe higher improvement gains on our LoRA finetuned version. 

\begin{table*}[t]
\centering
\caption{T2V CompBench results averaged across 5 runs for seven evaluation categories. Best scores within each model group are in \textbf{bold}. The Avg.\ column shows relative change vs.\ the group’s T2V baseline in parentheses. We include the best-performing model on the benchmark, the proprietary PixVerse-V3 for reference.} 
\label{tbl:compbench}
\scriptsize
\setlength{\tabcolsep}{4pt}
\renewcommand{\arraystretch}{1.1}
\resizebox{\textwidth}{!}{
\begin{tabular}{llccccccccc}
\toprule
\textbf{Model} & \textbf{Mode} & 
\begin{tabular}[c]{@{}c@{}}Consistent \\ Attribute\end{tabular} & 
\begin{tabular}[c]{@{}c@{}}Dynamic \\ Attribute\end{tabular} & 
\begin{tabular}[c]{@{}c@{}}Spatial \\ Relationship\end{tabular} & 
\begin{tabular}[c]{@{}c@{}}Motion \\ Binding\end{tabular} & 
\begin{tabular}[c]{@{}c@{}}Action \\ Binding\end{tabular} & 
\begin{tabular}[c]{@{}c@{}}Object \\ Interaction\end{tabular} & 
\begin{tabular}[c]{@{}c@{}}Generative \\ Numeracy\end{tabular} &
\textbf{Avg.} \\
\midrule
\textbf{SOTA (PixVerse-V3)} & --- & 0.706 & 0.062 & 0.598 & 0.287 & 0.872 & 0.831 & 0.607 & 0.566 \\
\midrule
\multirow{3}{*}{\textbf{WAN 2.2 (5B) ~\cite{wan2025wan}} } 
& T2V & 0.814 & 0.177 & 0.577 & 0.257 & 0.487 & 0.629 & 0.419 & 0.480 \\
& T2V (upsampled) & 0.866 & 0.137 & 0.606 & 0.245 & 0.509 & 0.690 & 0.523 & 0.511 {\scriptsize(\textcolor{green!50!black}{+6.46\%})} \\
& anchored T2V (Ours)  & \textbf{0.943} & \textbf{0.393} & \textbf{0.710} & \textbf{0.282} & \textbf{0.831} & \textbf{0.873} & \textbf{0.740} & \textbf{0.682} {\scriptsize(\textcolor{green!50!black}{+41.98\%})} \\
\midrule
\multirow{4}{*}{\textbf{WAN 2.2 (14B) ~\cite{wan2025wan}}} 
& T2V & 0.844 & 0.127 & 0.639 & 0.299 & 0.624 & 0.765 & 0.553 & 0.550 \\
& T2V (upsampled) & 0.897 & 0.084 & 0.643 & 0.291 & 0.675 & 0.762 & 0.605 & 0.565 {\scriptsize(\textcolor{green!50!black}{+2.73\%})} \\
& Anchored I2V + Text & 0.927 & \textbf{0.217} & 0.680 & 0.320 & \textbf{0.850} & 0.886 & \textbf{0.781} & \textbf{0.666} {\scriptsize(\textcolor{green!50!black}{+21.09\%})} \\
& Anchored T2V (Ours)  & \textbf{0.929} & 0.183 & \textbf{0.696} & \textbf{0.328} & 0.839 & \textbf{0.891} & 0.757 & 0.661 {\scriptsize(\textcolor{green!50!black}{+20.18\%})} \\
\midrule
\multirow{3}{*}{\textbf{CogVideo 1.5 (5B) ~\cite{hong2022cogvideo}}} 
& T2V & 0.762 & 0.191 & 0.491 & 0.252 & 0.308 & 0.491 & 0.305 & 0.400 \\
& T2V (upsampled) & 0.794 & 0.144 & 0.526 & 0.240 & 0.414 & 0.573 & 0.338 & 0.433 {\scriptsize(\textcolor{green!50!black}{+8.25\%})} \\
& Anchored T2V (Ours)  & \textbf{0.916} & \textbf{0.316} & \textbf{0.645} & \textbf{0.255} & \textbf{0.770} & \textbf{0.838} & \textbf{0.550} & \textbf{0.613} {\scriptsize(\textcolor{green!50!black}{+53.25\%})} \\
\midrule
\multirow{3}{*}{\textbf{WAN 2.1 (1B) ~\cite{wan2025wan}} } 
& T2V & 0.793 & 0.150 & 0.541 & 0.232 & 0.461 & 0.648 & 0.383 & 0.458 \\
& T2V (upsampled) & 0.871 & 0.117 & 0.590 & 0.244 & 0.572 & 0.684 & \textbf{0.470} & 0.507 {\scriptsize(\textcolor{green!50!black}{+10.72\%})} \\
& Anchored T2V (Ours) & \textbf{0.920} & \textbf{0.384} & \textbf{0.656} & \textbf{0.262} & \textbf{0.790} & \textbf{0.844} & 0.402 & \textbf{0.608} {\scriptsize(\textcolor{green!50!black}{+32.75\%})} \\
\bottomrule
\end{tabular}}
\end{table*}

\noindent \textbf{Results on VBench2.0:}
We summarize in~\cref{tbl:vbench} the results on VBench 2.0, reporting the five aggregated metrics. The complete set of 18 underlying metrics is provided in the Appendix. We report only the composition score instead of the original creativity metric (the latter averages composition and diversity) and study the diversity specifically in~\cref{sec:exp-diversity}.
Across all architectures, anchored T2V consistently improves every VBench metric except human fidelity, which shows a small decline even when using prompt upsampling. Similar to the observations on T2V-CompBench, our anchored Wan 5B outperforms the larger Wan 14B model, demonstrating that visual grounding is sometimes more beneficial than scaling.
While performance gains are consistent, improvements on VBench are generally smaller than on T2V-CompBench. This is expected since VBench applies a stricter evaluation protocol as most metrics are binary (0 or 1 per video) with no partial credit. Nonetheless, the trend confirms that visual anchoring systematically enhances compositional reasoning and controllability even under stricter assessment criteria.

\begin{table*}[t]
\centering
\caption{VBench 2.0 results averaged over five random seeds. We report four main evaluation categories, each aggregating 17 underlying metrics (see Appendix for the full breakdown). For reference, we also include results from the best-performing model on the benchmark, the proprietary Veo 3.}
\label{tbl:vbench}
\scriptsize
\setlength{\tabcolsep}{4pt}
\renewcommand{\arraystretch}{1.15}

\resizebox{\textwidth}{!}{
\begin{tabular}{llcccccc}
\toprule
\textbf{Model} & \textbf{Mode} & \textbf{Composition} & \textbf{Commonsense} & \textbf{Controllability} & \textbf{Human Fidelity} & \textbf{Physics} & \textbf{Avg.} \\
\midrule
\textbf{SOTA (Veo 3)} & --- & 0.696 & 0.695 & 0.470 & 0.869 & 0.693 & 0.685 \\
\midrule
\multirow{3}{*}{\textbf{WAN 2.2 (5B) ~\cite{wan2025wan}}}
& T2V & 0.488 & 0.635 & 0.268 & 0.826 & 0.527 & 0.549 \\
& T2V (upsampled) & 0.543 {\scriptsize(\textcolor{green!50!black}{+11.3\%})} & 0.594 {\scriptsize(\textcolor{red}{–6.5\%})} & 0.386 {\scriptsize(\textcolor{green!50!black}{+43.7\%})} & 0.806 {\scriptsize(\textcolor{red}{–2.5\%})} & 0.647 {\scriptsize(\textcolor{green!50!black}{+22.6\%})} & 0.595 {\scriptsize(\textcolor{green!50!black}{+8.4\%})} \\
& Anchored T2V (Ours) & 0.654 {\scriptsize(\textcolor{green!50!black}{+34.0\%})} & 0.647 {\scriptsize(\textcolor{green!50!black}{+1.8\%})} & 0.446 {\scriptsize(\textcolor{green!50!black}{+66.3\%})} & 0.780 {\scriptsize(\textcolor{red}{–5.6\%})} & 0.638 {\scriptsize(\textcolor{green!50!black}{+21.0\%})} & 0.633 {\scriptsize(\textcolor{green!50!black}{+15.3\%})} \\
\midrule
\multirow{4}{*}{\textbf{WAN 2.2 (14B) ~\cite{wan2025wan}}}
& T2V & 0.538 & 0.631 & 0.361 & 0.824 & 0.513 & 0.573 \\
& T2V (upsampled) & 0.557 {\scriptsize(\textcolor{green!50!black}{+3.6\%})} & 0.630 {\scriptsize(\textcolor{red}{–0.1\%})} & 0.434 {\scriptsize(\textcolor{green!50!black}{+20.3\%})} & 0.812 {\scriptsize(\textcolor{red}{–1.5\%})} & 0.670 {\scriptsize(\textcolor{green!50!black}{+30.7\%})} & 0.621 {\scriptsize(\textcolor{green!50!black}{+8.3\%})} \\
& Anchored T2V (Ours) & 0.708 {\scriptsize(\textcolor{green!50!black}{+31.6\%})} & 0.645 {\scriptsize(\textcolor{green!50!black}{+2.2\%})} & 0.473 {\scriptsize(\textcolor{green!50!black}{+31.1\%})} & 0.721 {\scriptsize(\textcolor{red}{–12.5\%})} & 0.685 {\scriptsize(\textcolor{green!50!black}{+33.7\%})} & 0.646 {\scriptsize(\textcolor{green!50!black}{+12.8\%})} \\
\midrule
\multirow{3}{*}{\textbf{CogVideo 1.5 (5B) ~\cite{hong2022cogvideo}}}
& T2V & 0.465 & 0.506 & 0.180 & 0.758 & 0.446 & 0.471 \\
& T2V (upsampled) & 0.479 {\scriptsize(\textcolor{green!50!black}{+2.8\%})} & 0.579 {\scriptsize(\textcolor{green!50!black}{+14.3\%})} & 0.198 {\scriptsize(\textcolor{green!50!black}{+10.2\%})} & 0.793 {\scriptsize(\textcolor{green!50!black}{+4.5\%})} & 0.531 {\scriptsize(\textcolor{green!50!black}{+18.9\%})} & 0.516 {\scriptsize(\textcolor{green!50!black}{+9.4\%})} \\
& Anchored T2V (Ours) & 0.649 {\scriptsize(\textcolor{green!50!black}{+39.4\%})} & 0.636 {\scriptsize(\textcolor{green!50!black}{+25.6\%})} & 0.304 {\scriptsize(\textcolor{green!50!black}{+69.3\%})} & 0.652 {\scriptsize(\textcolor{red}{–14.0\%})} & 0.562 {\scriptsize(\textcolor{green!50!black}{+25.8\%})} & 0.561 {\scriptsize(\textcolor{green!50!black}{+19.0\%})} \\
\bottomrule
\end{tabular}
}
\end{table*}

\begin{figure*}[t]
\label{fig:pull}
\centering
\includegraphics[width=\linewidth]{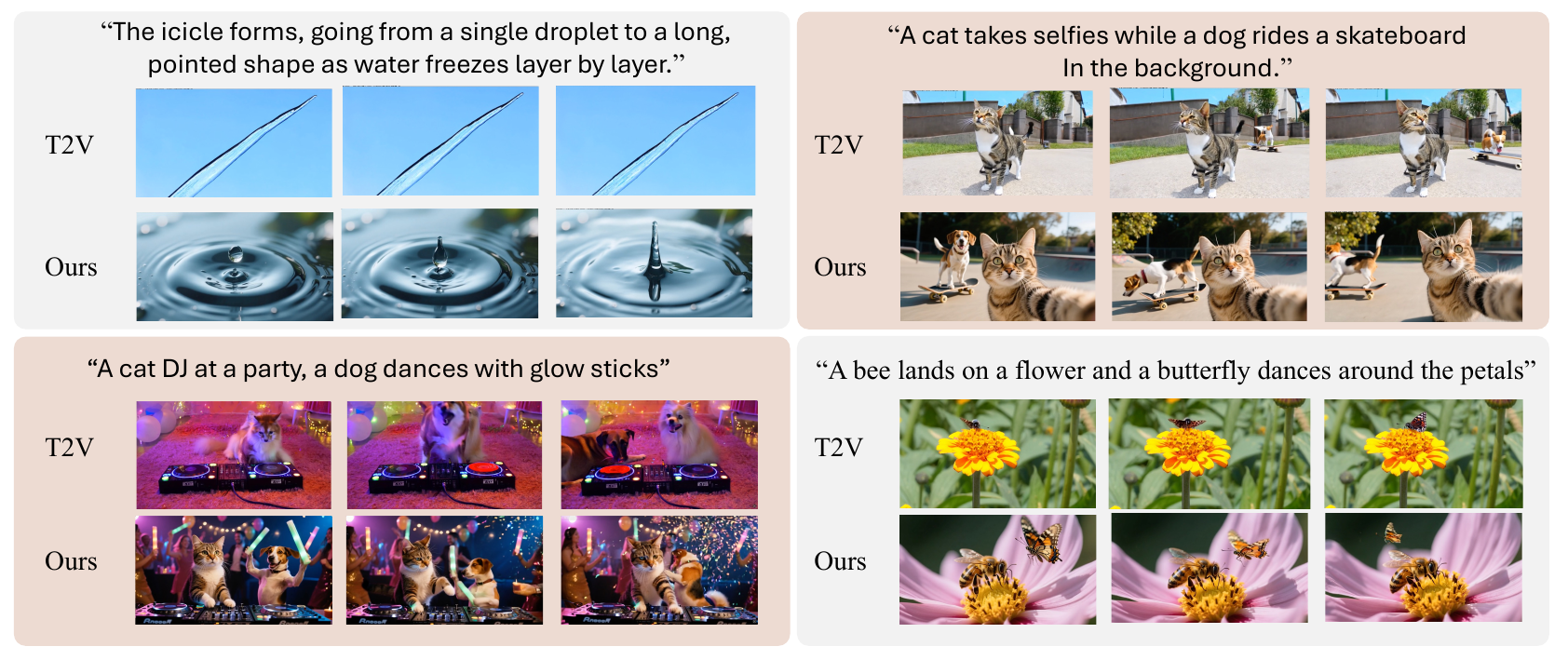}
\caption{Qualitative results showing our anchored method leads to better performance. Additional results are provided in the appendix.}
\label{fig:qualitative}
\end{figure*}

\noindent \textbf{Qualitative Results.}
\Cref{fig:qualitative} shows representative examples comparing text-only generation with our anchored approach. Anchored videos consistently exhibit more accurate scene composition, stronger object–attribute binding, and clearer temporal progression.

\subsection{Robustness to Reduced Sampling Steps}

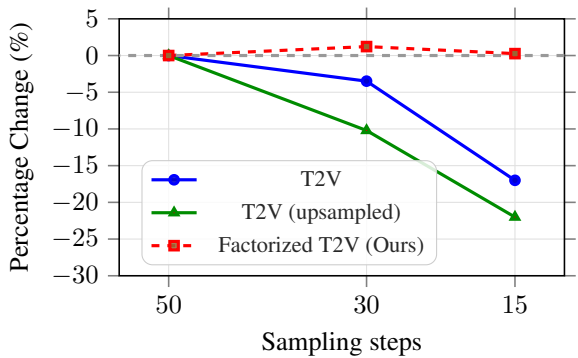
\begin{figure}[t]
\centering
\begin{tikzpicture}
\begin{axis}[
    width=0.7\linewidth,
    height=5cm,
    xmin=10, xmax=55,
    x dir=reverse,
    ymin=-30, ymax=5,
    xtick={50,30,15},
    ytick={-30,-25,-20,-15,-10,-5,0,5},
    yticklabel style={/pgf/number format/fixed},
    grid=both,
    grid style={draw=black!12},
    axis line style={semithick},
    tick style={semithick}, tick align=outside,
    xlabel={Sampling steps},
    ylabel={Percentage Change (\%)},
    legend columns=1,
    legend style={
        at={(0.05,0.45)},      % moved DOWN so it's inside the box
        anchor=north west,
        draw=black!20,
        rounded corners,
        fill=white, fill opacity=0.85,
        nodes={scale=0.85, transform shape},
        column sep=6pt,
        row sep=2pt
    },
    every axis plot/.append style={line width=1.2pt, mark options={solid, scale=0.8}}
]

% T2V — blue
\addplot+[blue, mark=*] coordinates {
    (50, 0)
    (30, -3.49)
    (15, -17.00)
};
\addlegendentry{T2V}

% T2V (upsampled) — green
\addplot+[green!55!black, mark=triangle*] coordinates {
    (50, 0)
    (30, -10.21)
    (15, -22.01)
};
\addlegendentry{T2V (upsampled)}

% Factorized T2V (Ours) — red
\addplot+[red, dashed, mark=square*] coordinates {
    (50, 0)
    (30, 1.22)
    (15, 0.26)
};
\addlegendentry{Anchored T2V (Ours)}

% Zero line
\addplot[black!40, dashed] coordinates {(10,0) (55,0)};
\node[anchor=south east, font=\scriptsize] at (axis cs:55,0.5) {Baseline (50 steps)};

\end{axis}
\end{tikzpicture}

\caption{Percentage change in WAN 2.2(5B) performance relative to 50-step baseline. Both T2V variants degrade as steps decrease, while ours remains stable at 15 steps.}
\label{fig:steps}
\end{figure}

We assess robustness to reduced diffusion steps by decreasing sampling from 50 to 30 and 15 for Wan5B and calculate the results on T2V Compbench based on 5 different seeds. As shown in~\cref{fig:steps}, the text-only model degrades sharply: –3.5\% at 30 steps and –17\% at 15 steps. The upsampled variant follows a similar trend but appears even more unstable (–10.2\% and –22.0\%), largely because it starts from a higher baseline at 50 steps. In contrast, the anchored model remains effectively unchanged, with only +1.2\% and +0.3\% variation. Although reducing diffusion from 50 to 15 steps would theoretically yield a 3.3× speedup, the full end-to-end pipeline—including anchor-image generation remains 2.1× faster in wall-clock time (1’39 vs.\ 3’30). Despite this substantial reduction in compute, the anchored model preserves the same accuracy, indicating that stronger conditioning not only improves quality but also stabilizes the diffusion trajectory, enabling far more efficient sampling.

% This means the anchored model maintains the same performance while using only one-third of the sampling steps, enabling nearly \textbf{4× faster} generation. These results suggest that stronger conditioning not only improves quality but also increases sampling efficiency by stabilizing the diffusion trajectory.

% once the scene is visually grounded, the diffusion model’s remaining task is largely temporal propagation, modeling how the world evolves, rather the dual task of scene reconstruction and temporal reasoning. 
% \begin{center}
% \colorbox{gray!10}{
% \parbox{0.95\linewidth}{
% \centering
% Visual grounding improves robustness and efficiency, maintaining performance with one-third of the steps.
% }}
% \end{center}

\subsection{Impact on Diversity} \label{sec:exp-diversity}
We evaluate generation diversity under several experimental settings, with results reported in \cref{tbl:diversity}. For each variation we generate 25 videos per prompt and calculate the diversity metric introduced by Vbench2.0.
The standard diversity evaluation involves fixing a text prompt and varying the seed. For a standard T2V model, this generates multiple videos from the same text. However, in our pipeline, a single seed change affects both the anchor image generation (T2I stage) and the subsequent video diffusion model. To disentangle these two sources of variance, we define two settings. First, we vary the seed for the entire pipeline. This captures the total diversity (variation in both scene and motion). This is reported as ``Seeds" in \cref{tbl:diversity}. Second, we generate a single anchor image, fix it, and then vary the seed only for the video diffusion model. This isolates motion-level diversity. This is reported as ``Single Image". We observe that our full pipeline yields diversity scores comparable to the baseline T2V model. Conversely, and as expected, fixing the anchor image and varying only the video model seed causes the diversity score to drop significantly. These experiments suggest that generation diversity could be usefully disentangled into scene-level diversity, which dominates the total variance, and motion-level diversity, which represents variations in animation for a static scene.
Finally, we evaluate an additional diversity enhancement strategy by using an LLM to rephrase prompts. For each original prompt, we generate 25 semantic variations and provide them to the models. This ``Prompt Rephrasing" method, as shown in \cref{tbl:diversity}, significantly boosts diversity scores across all tested models, confirming that semantic-level variation is a potent source of diverse outputs.

\begin{table}[t]
\centering
\caption{Diversity evaluation when varying seeds and prompts.}
\label{tbl:diversity}

\resizebox{\linewidth}{!}{%
\setlength{\tabcolsep}{4pt}
\renewcommand{\arraystretch}{1.1}
\begin{tabular}{lcccc}
\toprule
\begin{tabular}[c]{@{}c@{}}\textbf{Model} \\ \textbf{Variable}\end{tabular}
&
\begin{tabular}[c]{@{}c@{}}\textbf{T2V} \\ \textbf{Seeds}\end{tabular} &
\begin{tabular}[c]{@{}c@{}}\textbf{T2V} \\ \textbf{Prompt Rephrasing}\end{tabular} &
\begin{tabular}[c]{@{}c@{}}\textbf{Anchored T2V (Ours)} \\ \textbf{Single Image}\end{tabular} &
\begin{tabular}[c]{@{}c@{}}\textbf{Anchored T2V (Ours)} \\ \textbf{Seeds}\end{tabular} \\
\midrule
\textbf{WAN 5B}  & 0.578 & 0.816 & 0.2404 & 0.539 \\
\textbf{WAN 14B} & 0.531 & 0.742 & 0.348  & 0.513 \\
\bottomrule
\end{tabular}%
}
\vspace{-8mm}
\end{table}

\section{Ablation Studies}
To better understand the components contributing to the effectiveness of our framework, we conduct three ablations. First, we study the flexibility of anchor placement by evaluating anchoring at different temporal positions.  Second, we analyze the impact of LLM-based prompt rewriting for anchor image generation. Third, we evaluate the model's sensitivity to different anchor images.

% \subsection{Flexible Anchor Position}
% We study whether AVG performance depends on first-frame anchoring, or whether anchors can be placed at other temporal positions. To test this, we modify the reasoning stage to predict the middle-frame scene state rather than the initial one, and use this temporally aligned description for composition and anchoring. This is enabled by our anchor-grounding finetuning, which trains the model to interpret a clean injected frame as a valid anchor at arbitrary positions. Under this setting, middle-frame anchoring remains comparable to first-frame anchoring, achieving 0.670 vs.\ 0.661 on CompBench and 0.631 vs.\ 0.646 on VBench2. These results suggest that the exact anchor location is not the key factor; what matters is producing the correct visual state at the chosen temporal position. Once the scene is properly grounded, temporal synthesis becomes significantly easier, as the model can focus on motion rather than jointly solving scene construction and animation. Overall, this highlights AVG as a flexible pipeline that can incorporate anchors at different temporal locations.

\subsection{Flexible Anchor Position}

\begin{wraptable}{r}{0.38\linewidth}
\vspace{-2mm}
\centering
\scriptsize
\setlength{\tabcolsep}{4pt}
\renewcommand{\arraystretch}{1.05}
\begin{tabular}{lcc}
\toprule
\textbf{Anchor} & \textbf{T2VComp.} & \textbf{VB2} \\
\midrule
First  & 0.661 & 0.646 \\
Middle & 0.670 & 0.631 \\
\bottomrule
\end{tabular}
\caption{AVG Wan14B with different anchor positions.}
\label{tab:anchor_position}
\vspace{-4mm}
\end{wraptable}

We study whether AVG depends on first-frame anchoring, or whether anchors can be placed at other temporal positions. To test this, we modify the reasoning stage to predict the middle-frame scene state and use this temporally aligned description for composition and anchoring. This is enabled by our anchor-grounding finetuning, which allows a clean injected frame to serve as a valid anchor at arbitrary positions. As shown in Table~\ref{tab:anchor_position}, middle-frame anchoring remains comparable to first-frame anchoring. This suggests that the exact anchor location is not the key factor; what matters is producing the correct visual state at the chosen temporal position. Once the scene is properly grounded, temporal synthesis becomes easier, as the model can focus on motion rather than jointly solving scene construction and animation. Overall, this highlights AVG as a flexible pipeline that can incorporate anchors at different temporal locations.

\subsection{Original Caption vs. LLM-based First-Frame Scene Description}
\label{ablation:caption}

We examine whether rewriting the video prompt using an LLM produces a more temporally appropriate anchor image—i.e., an image that correctly represents the scene as it should appear in the first frame. Specifically, we compare two settings on \textsc{Wan 5B}: (1) using the raw video caption directly as the image-generation prompt, and (2) using our LLM-refined first-frame description. On VBench 2.0, the direct-caption baseline achieves an average score of 55.8\% which is to be compared to the T2V baseline of 54.9\%. Our LLM-guided anchoring on the contrary reaches 63.3\%, corresponding to a substantial relative improvement of $15.3\%$. These results highlight the importance of using an LLM to reason and generate a proper scene description. Providing a semantically accurate and temporally plausible starting scene is crucial. In contrast, using an unrefined caption often produces a mid-video moment, leading to weaker grounding and significantly lower overall performance (see qualitatives examples in Appendix).

\subsection{Sensitivity to the Choice of Anchor Image} \label{sec:ablation-perf}

While our main experiments use a single preferred anchor image, mimicking a human-in-the-loop workflow, we also evaluate how sensitive the model is to the specific anchor provided. For Wan 5B, we generate five different anchor images per prompt using our LLM-refined first-frame description and assess the resulting videos across all VBench 2.0 categories. As shown in~\cref{tbl:anchor-seed-std}, the variation across different image anchors is extremely small: all categories have standard deviations below 0.02. This indicates that the slight variations resulting from the stochastic generation of anchor images affect mostly appearance and not the model's compositional performance. 
To contextualize this robustness, \cref{tbl:anchor-seed-std} compares the variability of our anchored model across anchors with the variability of the original T2V model across random seeds. On average, T2V is \textbf{9× more variable} with respect to seed than with respect to image anchors. This indicates the effectiveness of our anchored approach.
\begin{table}[t]
\centering
\caption{Standard deviation across anchor images (Anchored Wan5B) vs.\ across seeds (Wan5B). Lower values indicate greater stability.}
\label{tbl:anchor-seed-std}
\scriptsize
\setlength{\tabcolsep}{6pt}
\renewcommand{\arraystretch}{1.1}

\begin{tabular}{lcc}
\toprule
\textbf{Category} &
\begin{tabular}[c]{@{}c@{}} \textbf{Anchored Wan5B (Ours)} \\ \textbf{(Std. Anchors)} \end{tabular} &
\begin{tabular}[c]{@{}c@{}} \textbf{Wan5B} \\ \textbf{(Std. Seeds)} \end{tabular} \\
\midrule
Consistent Attribute   & 0.0039 & 0.0117 \\
Dynamic Attribute      & 0.0113 & 0.0203 \\
Spatial Relationship   & 0.0092 & 0.1659 \\
Motion Binding         & 0.0031 & 0.0053 \\
Action Binding         & 0.0144 & 0.2180 \\
Object Interaction     & 0.0128 & 0.2401 \\
Generative Numeracy    & 0.0201 & 0.0171 \\
\midrule
\textbf{Average}       & 0.0107 & 0.0969 \\
\bottomrule
\end{tabular}
\end{table}

%% file: sec/5_discussion.tex
\section{Discussion}
\noindent\textbf{Anchoring  vs. model scaling.}
Our results suggest that improved grounding, rather than increased capacity alone, may be equally important for T2V generation. Recent advances in T2V diffusion have relied heavily on scaling model size and training data, yet even large models often struggle to infer a coherent initial scene from text alone. This contrasts with image diffusion, where scaling is relatively straightforward; in video models, each architectural improvement must operate over an additional temporal dimension, making scaling substantially more resource intensive. Our findings indicate that stronger visual grounding can complement scale by addressing a different bottleneck: establishing the correct scene before motion synthesis. By separating scene composition from temporal modeling through a visual anchor, we mitigate several common failure modes without requiring substantially larger models. We view this as a complementary design principle for building more reliable and structured video models.

\noindent\textbf{Implications for T2V evaluation.}
The effectiveness of visual anchoring also highlights a potential gap in current evaluation practice. Benchmarks such as VBench2.0 and T2V-CompBench reward accurate scene composition and prompt adherence, yet we find that standard T2V models often underperform relative to a simple anchored baseline. This suggests that future evaluations may benefit from including anchor-conditioned or native I2V+text baselines to better disentangle scene construction from true temporal modeling. More broadly, if the goal is to measure progress in motion understanding rather than improvements in static composition alone, it may also be valuable to evaluate anchored counterparts of T2V models under the same backbone.

\noindent\textbf{Performance on Human Fidelity.}
As shown in~\cref{tbl:vbench}, both prompt upsampling and visual anchoring tend to reduce performance on the Human Fidelity category. This aggregate score combines three sub-metrics: human anatomy, human clothes, and human identity. As detailed in the full results in~\cref{tbl:vbench_full} (Appendix), the decline is driven almost entirely by the human identity component, while the other two components remain stable. Through manual inspection, we observe that all model variants, regardless of anchoring, struggle to preserve identity especially across different actions. 
While our primary focus is improving compositional and reasoning capabilities, identity preservation is a distinct and challenging problem in its own. Addressing this limitation, potentially through explicit identity conditioning or persistent character representations, remains an important direction for future work.

\noindent\textbf{On the diversity--correctness trade-off.}
Diversity is important for applications such as synthetic data generation and counterfactuals, where multiple distinct videos must be produced without violating the intended scenario. However, current diversity metrics are based on VGG feature differences and mainly capture scene appearance. As shown in our performance ablation in~\cref{sec:ablation-perf}, variation across seeds can significantly affect correctness, so these metrics may count compositional failures or inconsistent layouts as ``diversity.'' Our anchored pipeline reduces prompt-adherence failures, as reflected in its stronger benchmark scores, but although anchoring improves correctness, it does not guarantee diversity under existing metrics. Conversely, standard T2V attains higher numerical diversity by exploring a broader but less controlled generative space, which can also yield unfaithful or incorrect outputs.
We therefore argue that a faithful diversity metric should measure not only whether outputs vary, but also whether they satisfy the intended constraints. More specifically, true video diversity should capture two dimensions: (1) appearance diversity across frames, which current metrics partially measure, and (2) motion diversity under fixed conditions. A principled way to evaluate motion diversity may be to measure variability in an anchored model's outputs while keeping the first frame fixed, thereby isolating motion from scene composition.

% \noindent\textbf{On the diversity–correctness trade-off.}
% Diversity is essential for applications such as synthetic data generation and counterfactuals, where one must produce multiple distinct videos without violating the intended scenario. Yet current diversity metrics rely on VGG feature differences, which primarily capture scene appearance. As shown in our performance ablation in~\cref{sec:ablation-perf}, variability across seeds can meaningfully affect correctness, meaning that these metrics may interpret clear compositional failures or inconsistent layouts as “diversity". Our anchored pipeline reduces prompt-adherence failures, as evidenced by its stronger benchmark scores, but while anchoring improves correctness, it does not guarantee diversity under current metrics. Conversely, standard T2V achieves higher numerical diversity but explores a broader but less controlled generative space, sometimes producing unfaithful or incorrect outcomes.
% We argue that the introduction of faithful diversity metric ensuring not only that the outputs vary but also that they respect certain constraints would be highly valuable.
% We also suggest that true video diversity should reflect two distinct diversity dimensions. First, the appearance diversity of the frames, which is captured with current metrics. Second, the motion diversity under given conditions. We argue that a possible principled way to evaluate motion diversity is to quantify the variability in a anchored model’s outputs when the first frame is kept constant, isolating motion from scene composition.

\section{Conclusion}
We proposed a simple anchored generation framework, a simple pipeline that grounds the first frame before synthesizing motion, addressing a key failure mode of current T2V models. This explicit visual anchoring yields strong gains on compositional and reasoning benchmarks, with small models outperforming larger baselines and maintaining accuracy at one-third the sampling steps. While challenges such as identity preservation remain, our results show that explicitly separating scene construction from temporal modeling is a powerful and practical direction for future video generation research.

%% file: sec/X_suppl.tex
\clearpage

\appendix
\begin{center}
\vspace{1em}
    {\begin{center}
    {\large  \bfseries Anchored Video Generation: Decoupling Scene Construction and Temporal Synthesis in Text-to-Video Diffusion Models\par}
    \vspace{0.75em}
    {\large Supplementary Material\par}
\end{center}}
\end{center}

% \setcounter{page}{1}
% \title{Anchored Video Generation: Decoupling Scene Construction and Temporal Synthesis in Text-to-Video Diffusion Models} 
% \maketitle

% \section{Video Examples}
% We include an additional folder, \texttt{Website}, containing an interactive HTML file that opens a webpage with extensive qualitative video comparisons between the standard T2V models and our Anchored T2V approach. These examples provide visual evidence of the qualitative trends reported in the main paper, including enhanced scene grounding and more reliable temporal evolution.

\section{FLops Analysis}
In this section, we give a more detailed study of the computation required by each model. The anchoredd version required a diffusion model for image generation and one for video generation. We report the Flops for a single call for each model. During sampling, the models are called multiple times for inference. 
As a video model, we report the Flops for Wan 2.2 with 5 billion parameters. For the image model, we report the Flops for Qwen-Image with 20 billion parameters. A single call of Wan 2.2 represents $117 \times 10^3$ GFLOPS. To generate a video, we use 50 inference steps and classifier-free guidance. As a result, the total cost of generating a video is $117 \times 10^5$ GFLOPS.
A single call of Qwen-Image represents $10.7$ GFlOPS. To generate an image, we use 50 inference steps and classifier-free guidance. As a result, generating an image costs about $10.7 \times 10^2$ GFLOPS. Finally, we use the anchored approach to reduce the number of inference steps from 50 to 15 for video generation. The total cost is $35.1 \times 10^5$ GFLOPS.

\section{Detailed Experiments Results}

\subsection{VBench 2.0}

\Cref{tbl:vbench_full} show the detailed metric results for VBench2.0 averaged across five seeds for Wan2.2 5B, Wan2.2 14B and CogVideo1.5 5B. Across the full set of 17 VBench2.0 metrics, we observe consistent trends that mirror the aggregated results in \cref{tbl:vbench}. First, anchored T2V improves nearly every metric relative to both T2V and upsampled T2V for all three model sizes (5B, 14B, and 5B CogVideoX). Improvements are especially significant in categories reflecting reasoning-driven scene understanding such as Composition, Instance Preservation,  Motion Order Understanding, Dynamic Attribute and Dynamic Spatial Relationship further supporting our claim that anchoring mitigates the dominant failure mode of T2V: incorrect scene construction.

Second, we find that WAN 5B with anchoring often outperform WAN 14B T2V, despite having far fewer parameters. This confirms that scaling alone does not guarantee robust compositional reasoning, and that visual grounding provides a stronger foundation.

Third, consistent with our observations in \cref{tbl:vbench}, anchoring does not improve performance on Human Identity. The identity-related metric drops even when using upsampled prompts, indicating that this limitation stems from the underlying generative model.

Overall, the detailed metric breakdown reinforces that anchoring strengthens video diffusion by correcting their weak scene construction and making use of their excellent temporal modeling which yields broad improvements in compositional correctness while remaining model and scale-agnostic.

\begin{table*}[t]
\centering
\caption{VBench 2.0 results averaged over five random seeds. We report raw per-metric scores, later aggregated into five categories.}
\label{tbl:vbench_full}
\scriptsize
\setlength{\tabcolsep}{3pt}
\renewcommand{\arraystretch}{1.08}

% ---------- Part 1: First set of metrics ----------
\resizebox{\textwidth}{!}{
\begin{tabular}{llcccccccccc}
\toprule
\textbf{Model} & \textbf{Mode} &
\begin{tabular}[c]{@{}c@{}}\textbf{Camera}\\ \textbf{Motion}\end{tabular} &
\begin{tabular}[c]{@{}c@{}}\textbf{Complex}\\ \textbf{Landscape}\end{tabular} &
\begin{tabular}[c]{@{}c@{}}\textbf{Complex}\\ \textbf{Plot}\end{tabular} &
\textbf{Composition} &
\begin{tabular}[c]{@{}c@{}}\textbf{Dynamic}\\ \textbf{Attribute}\end{tabular} &
\begin{tabular}[c]{@{}c@{}}\textbf{Dynamic Spatial}\\ \textbf{Relationship}\end{tabular} &
\begin{tabular}[c]{@{}c@{}}\textbf{Human}\\ \textbf{Anatomy}\end{tabular} &
\begin{tabular}[c]{@{}c@{}}\textbf{Human}\\ \textbf{Clothes}\end{tabular} &
\begin{tabular}[c]{@{}c@{}}\textbf{Human}\\ \textbf{Identity}\end{tabular} \\
\midrule
\textbf{SOTA (Veo3)} & --- &
0.5494 & 0.2178 & 0.2187 & 0.6857 & 0.6374 & 0.4589 & 0.9026 & 0.9953 & 0.7084 \\
\midrule
\multirow{3}{*}{\textbf{WAN 5B}} 
& T2V &
0.1630 & 0.1960 & 0.1156 & 0.4880 & 0.1319 & 0.3594 & 0.8811 & 0.9150 & 0.6826 \\
& T2V (upsampled) &
0.4963 & 0.2213 & 0.1279 & 0.5433 & 0.4066 & 0.3246 & 0.8763 & 0.9284 & 0.6133 \\
& Anchored T2V &
0.5685 & 0.2187 & 0.1677 & 0.6537 & 0.5670 & 0.4232 & 0.8904 & 0.8526 & 0.5967 \\
\midrule
\multirow{3}{*}{\textbf{WAN 14B}} 
& T2V &
0.2148 & 0.1967 & 0.1443 & 0.5377 & 0.4872 & 0.4203 & 0.9266 & 0.8531 & 0.6919 \\
& T2V (upsampled) &
0.5407 & 0.1960 & 0.1592 & 0.5570 & 0.5659 & 0.3246 & 0.9294 & 0.9549 & 0.5509 \\
& Anchored T2V &
0.5741 & 0.2367 & 0.1796 & 0.7077 & 0.7473 & 0.3768 & 0.9310 & 0.8663 & 0.3653 \\
\midrule
\multirow{3}{*}{\textbf{CogVideoX 5B}} 
& T2V &
0.2500 & 0.1880 & 0.0353 & 0.4655 & 0.1055 & 0.1507 & 0.5895 & 0.9050 & 0.7803 \\
& T2V (upsampled) &
0.2407 & 0.1693 & 0.0626 & 0.4786 & 0.1231 & 0.1797 & 0.7215 & 0.8371 & 0.8190 \\
& Anchored T2V &
0.2722 & 0.2227 & 0.1483 & 0.6490 & 0.2813 & 0.2754 & 0.7843 & 0.6306 & 0.5410 \\
\bottomrule
\end{tabular}
}

\vspace{0.4em}

% ---------- Part 2: Remaining metrics ----------
\resizebox{\textwidth}{!}{
\begin{tabular}{llcccccccc}
\toprule
\textbf{Model} & \textbf{Mode} &
\begin{tabular}[c]{@{}c@{}}\textbf{Human}\\ \textbf{Interaction}\end{tabular} &
\begin{tabular}[c]{@{}c@{}}\textbf{Instance}\\ \textbf{Preservation}\end{tabular} &
\textbf{Material} &
\textbf{Mechanics} &
\begin{tabular}[c]{@{}c@{}}\textbf{Motion Order}\\ \textbf{Understanding}\end{tabular} &
\begin{tabular}[c]{@{}c@{}}\textbf{Motion}\\ \textbf{Rationality}\end{tabular} &
\begin{tabular}[c]{@{}c@{}}\textbf{Multi-View}\\ \textbf{Consistency}\end{tabular} &
\textbf{Thermotics} \\
\midrule
\textbf{SOTA (Veo3)} & --- &
0.8067 & 0.9298 & 0.8411 & 0.8182 & 0.4040 & 0.4598 & 0.3595 & 0.7551 \\
\midrule
\multirow{3}{*}{\textbf{WAN 5B}} 
& T2V &
0.6680 & 0.8807 & 0.4031 & 0.5848 & 0.2449 & 0.3897 & 0.5235 & 0.5976 \\
& T2V (upsampled) &
0.7880 & 0.7880 & 0.7348 & 0.6890 & 0.3354 & 0.4000 & 0.4751 & 0.6873 \\
& Anchored T2V &
0.7940 & 0.8316 & 0.7813 & 0.7768 & 0.3859 & 0.4621 & 0.3811 & 0.6125 \\
\midrule
\multirow{3}{*}{\textbf{WAN 14B}} 
& T2V &
0.7675 & 0.8772 & 0.5916 & 0.5837 & 0.2929 & 0.3851 & 0.3025 & 0.5727 \\
& T2V (upsampled) &
0.8320 & 0.7851 & 0.8011 & 0.7464 & 0.4182 & 0.4759 & 0.4818 & 0.6513 \\
& Anchored T2V &
0.8100 & 0.8421 & 0.7436 & 0.8459 & 0.3838 & 0.4483 & 0.4354 & 0.7171 \\
\midrule
\multirow{3}{*}{\textbf{CogVideoX 5B}} 
& T2V &
0.4600 & 0.7404 & 0.4271 & 0.6742 & 0.0670 & 0.2724 & 0.0969 & 0.5873 \\
& T2V (upsampled) &
0.5300 & 0.7439 & 0.6237 & 0.6438 & 0.0793 & 0.4138 & 0.2608 & 0.5946 \\
& Anchored T2V &
0.7420 & 0.8070 & 0.7587 & 0.7970 & 0.1859 & 0.4655 & 0.0150 & 0.6761 \\
\bottomrule
\end{tabular}
}

\end{table*}

\subsection{Robustness to Reduced Sampling Steps}

Table \ref{tbl:robustness} and the full metric values reported in Table \ref{tbl:robustness-raw} show a clear difference in robustness of each model to a reduction in number of sampling steps. The text-only T2V model's performance drops at 30 steps (–3.5\% on average) and collapses at 15 steps (–17\%), with large drop in categories such as Action Binding and Object Interaction. The upsampled T2V variant follows a similar pattern. Due to the fact that it begins with a higher 50-step baseline, its average performance decreases by –10.2\% at 30 steps and –22.0\% at 15 steps, with some metrics falling by over –35\%. On the other hand, the anchored model remains remarkably stable. Even when the sampling steps are reduced to 30 or 15, its average score changes by only +1.22\% and +0.26\%, respectively. These results indicate that visual anchoring significantly improves robustness to aggressive step reduction. Thus, once the initial scene is grounded, generation relies far less on long denoising steps.

\begin{table*}[t]
\centering
\caption{Percentage change in WAN 5B performance when reducing sampling steps. The baseline results are at 50 sampling steps.}
\label{tbl:robustness}
\scriptsize
\setlength{\tabcolsep}{5pt}
\renewcommand{\arraystretch}{1.1}
\resizebox{\textwidth}{!}{
\begin{tabular}{llccccccccc}
\toprule
\textbf{Model} & \textbf{Mode} & \textbf{Steps} &
\textbf{Consistency} & \textbf{Dynamic} & \textbf{Spatial} &
\textbf{Motion} & \textbf{Action} & \textbf{Interaction} &
\textbf{Numeracy} & \textbf{Avg.} \\
\midrule
\multirow{6}{*}{\textbf{WAN 5B}}
& T2V  & 15 &
\textcolor{red}{-13.86} & \textcolor{red}{-14.62} & \textcolor{red}{-5.07} &
\textcolor{red}{-6.93} & \textcolor{red}{-31.80} & \textcolor{red}{-20.34} &
\textcolor{red}{-24.51} & \textcolor{red}{-17.00} \\
& T2V  & 30 &
\textcolor{red}{-2.44} & \textcolor{red}{-2.86} & \textcolor{red}{-0.13} &
\textcolor{green!50!black}{1.10} & \textcolor{red}{-8.03} &
\textcolor{red}{-3.54} & \textcolor{red}{-7.87} & \textcolor{red}{-3.49} \\
\addlinespace[2pt]
\cmidrule(lr){2-11}
& T2V (upsampled) & 15 &
\textcolor{red}{-19.00} & \textcolor{green!50!black}{10.72} &
\textcolor{red}{-9.74} & \textcolor{red}{-2.36} &
\textcolor{red}{-34.73} & \textcolor{red}{-27.38} &
\textcolor{red}{-39.48} & \textcolor{red}{-22.01} \\
& T2V (upsampled) & 30 &
\textcolor{red}{-8.71} & \textcolor{green!50!black}{12.48} &
\textcolor{red}{-4.99} & \textcolor{green!50!black}{5.66} &
\textcolor{red}{-14.01} & \textcolor{red}{-11.90} &
\textcolor{red}{-26.16} & \textcolor{red}{-10.21} \\
\addlinespace[2pt]
\cmidrule(lr){2-11}
& Anchored T2V  & 15 &
\textcolor{green!50!black}{0.26} & \textcolor{red}{-2.27} &
\textcolor{green!50!black}{0.18} & \textcolor{red}{-1.66} &
\textcolor{green!50!black}{0.03} & \textcolor{green!50!black}{2.96} &
\textcolor{red}{-0.50} & \textcolor{green!50!black}{0.26} \\
& Anchored T2V  & 30 &
\textcolor{green!50!black}{0.02} & \textcolor{green!50!black}{0.68} &
\textcolor{green!50!black}{0.16} & \textcolor{red}{-0.35} &
\textcolor{green!50!black}{1.65} & \textcolor{green!50!black}{3.66} &
\textcolor{green!50!black}{1.23} & \textcolor{green!50!black}{1.22} \\
\bottomrule
\end{tabular}}
\end{table*}

\begin{table*}[t]
\centering
\caption{Average WAN-5B results across five seeds for T2V, upsampled T2V, and our anchored T2V at different sampling steps. Our method remains stable even at 15 steps, whereas both T2V variants degrade sharply as steps decrease.}
\label{tbl:robustness-raw}
\scriptsize
\setlength{\tabcolsep}{5pt}
\renewcommand{\arraystretch}{1.1}
\resizebox{\textwidth}{!}{
\begin{tabular}{llccccccccc}
\toprule
\textbf{Model} & \textbf{Mode} & \textbf{Steps} &
\textbf{Consistency} & \textbf{Dynamic} & \textbf{Spatial} &
\textbf{Motion} & \textbf{Action} & \textbf{Interaction} &
\textbf{Numeracy} & \textbf{Avg.} \\
\midrule
\multirow{9}{*}{\textbf{WAN 5B}}
& T2V & 15 &
0.7013 & 0.1515 & 0.5480 & 0.2391 & 0.3323 & 0.5008 & 0.3167 & 0.3985 \\
& T2V & 30 &
0.7943 & 0.1723 & 0.5765 & 0.2597 & 0.4481 & 0.6065 & 0.3865 & 0.4634 \\
& T2V & 50 &
0.8142 & 0.1774 & 0.5773 & 0.2569 & 0.4872 & 0.6287 & 0.4195 & 0.4802 \\
\addlinespace[2pt]
\cmidrule(lr){2-11}
& T2V (upsampled) & 15 &
0.7013 & 0.1515 & 0.5468 & 0.2391 & 0.3323 & 0.5008 & 0.3167 & 0.3983 \\
& T2V (upsampled) & 30 &
0.7904 & 0.1539 & 0.5756 & 0.2587 & 0.4377 & 0.6075 & 0.3863 & 0.4586 \\
& T2V (upsampled) & 50 &
0.8658 & 0.1368 & 0.6059 & 0.2449 & 0.5090 & 0.6896 & 0.5232 & 0.5107 \\
\addlinespace[2pt]
\cmidrule(lr){2-11}
& Anchored T2V & 15 &
0.9399 & 0.3836 & 0.6606 & 0.2799 & 0.8214 & 0.8953 & 0.7512 & 0.6760 \\
& Anchored T2V & 30 &
0.9377 & 0.3951 & 0.6604 & 0.2836 & 0.8347 & 0.9013 & 0.7642 & 0.6824 \\
& Anchored T2V & 50 &
0.9375 & 0.3925 & 0.6594 & 0.2846 & 0.8212 & 0.8695 & 0.7550 & 0.6742 \\
\bottomrule
\end{tabular}}
\end{table*}

\section{Qualitative Results}

\begin{figure*}[t]
\label{fig:pull}
\centering
\includegraphics[width=\linewidth]{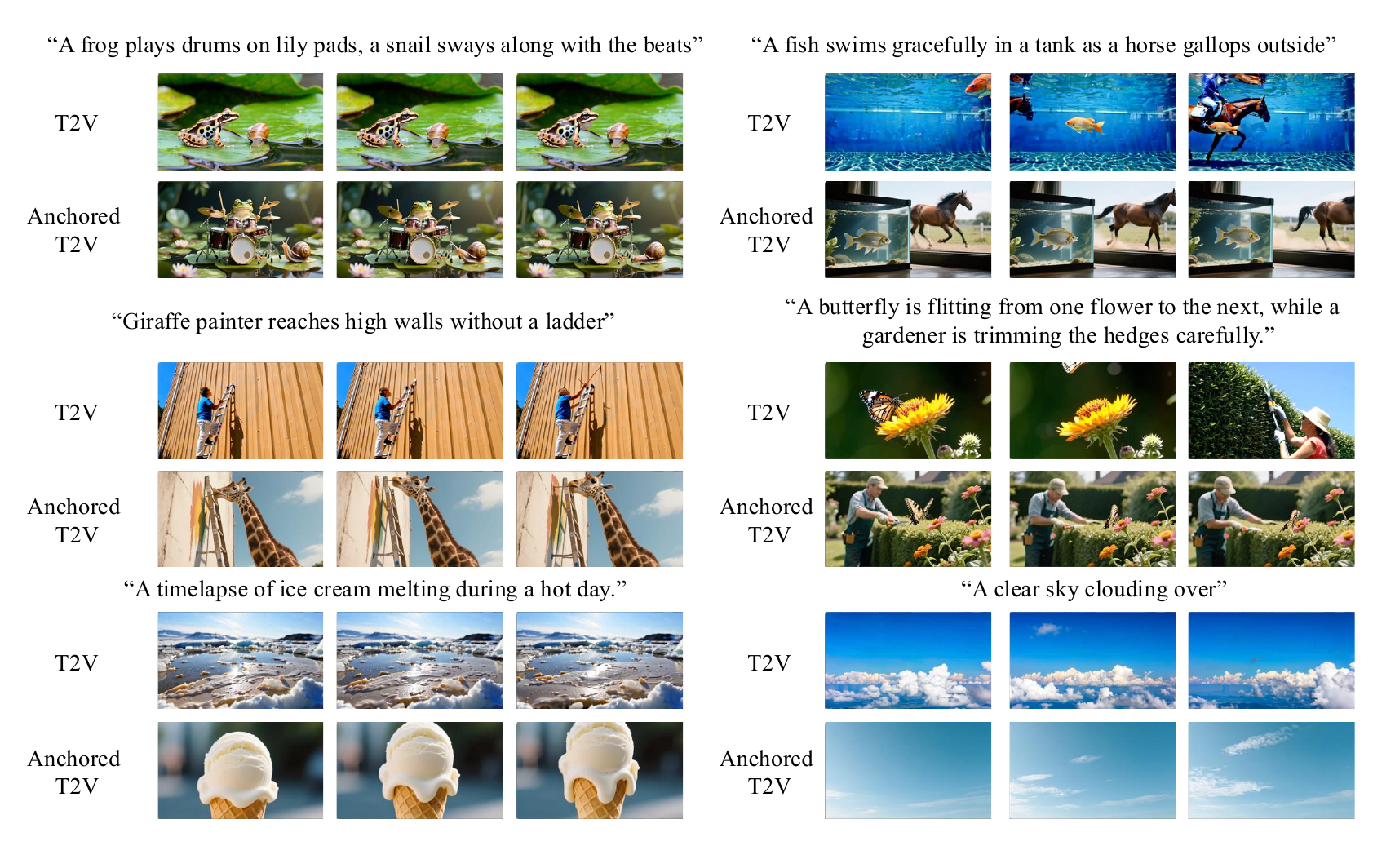}
\caption{Overall qualitative results comparing our anchored method against T2V methods. Visual examples from Wan14B showing that anchored T2V produces more coherent scene layouts and semantically aligned compositions than text-only T2V methods..}
\label{fig:qualitative}
\end{figure*}

\begin{figure*}[t]
\label{fig:pull}
\centering
\includegraphics[width=\linewidth]{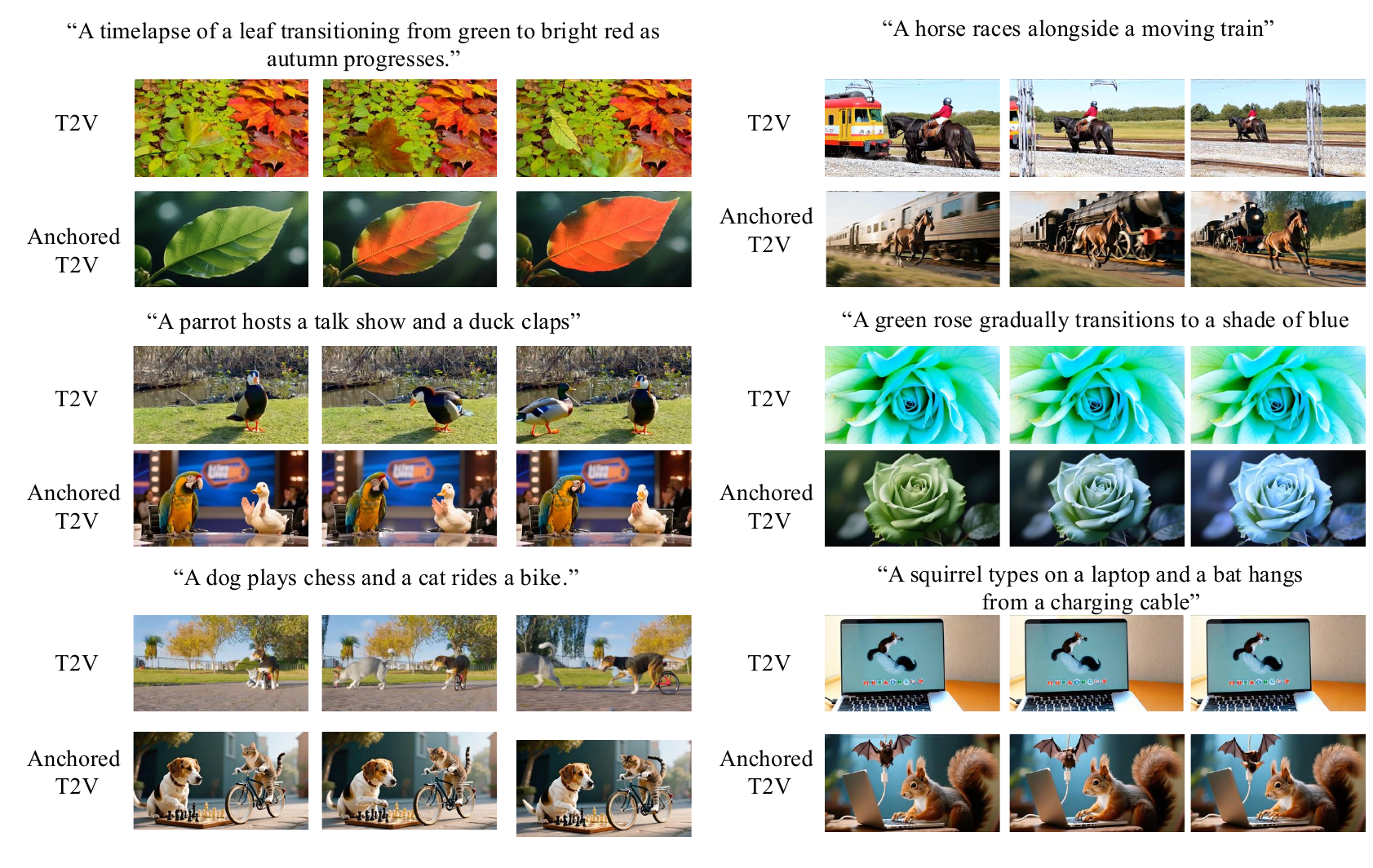}
\caption{Overall qualitative results comparing our anchored method against T2V methods. Visual examples from Wan5B showing that anchored T2V produces more coherent scene layouts and semantically aligned compositions than text-only T2V methods..}
\label{fig:wan5b}
\end{figure*}

\begin{figure*}[t]
\label{fig:pull}
\centering
\includegraphics[width=\linewidth]{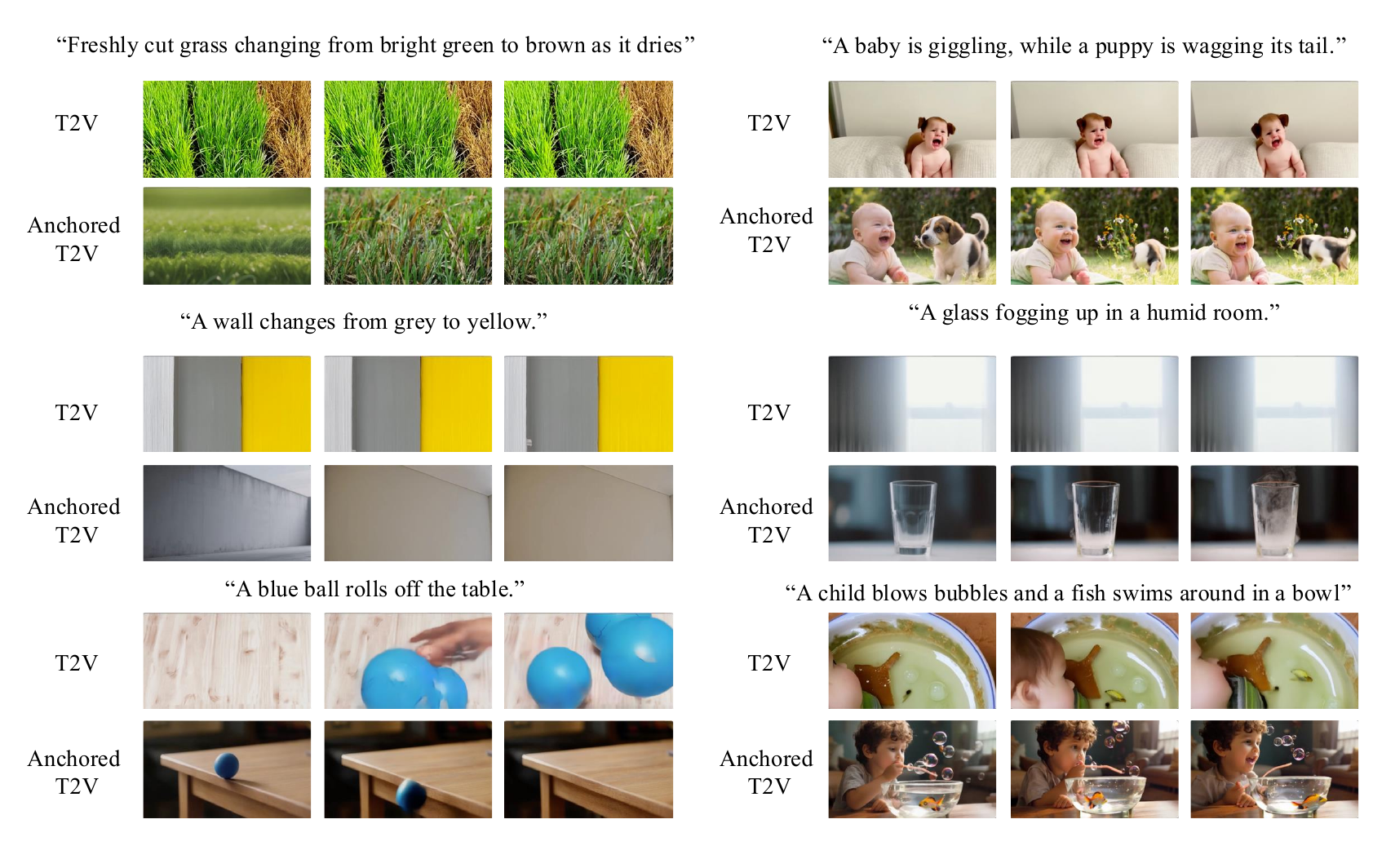}
\caption{Overall qualitative results comparing our anchored method against T2V methods. Visual examples from CogVideo showing that anchored T2V produces more coherent scene layouts and semantically aligned compositions than text-only T2V methods.}
\label{fig:cogvideo}
\end{figure*}

\paragraph{Anchored Scene Construction Across Multiple Video Models.}
\Cref{fig:qualitative}, \cref{fig:wan5b}, and  \cref{fig:cogvideo} provides additional examples illustrating the visual differences between standard T2V generation and our anchored approach for Wan14B, Wan5B and CogVideo5B respectively across a wide variety of scenes. Across diverse prompts, the anchored model produces scenes with more accurate semantic alignment with the described situation. On the other hand, text-only T2V baselines often hallucinate or misinterpret spatial relations, or produce semantically incorrect first frames. These examples complement our quantitative results and highlight how grounding the visual anchor substantially improves scene construction.

\begin{figure*}[t]
\label{fig:pull}
\centering
\includegraphics[width=\linewidth]{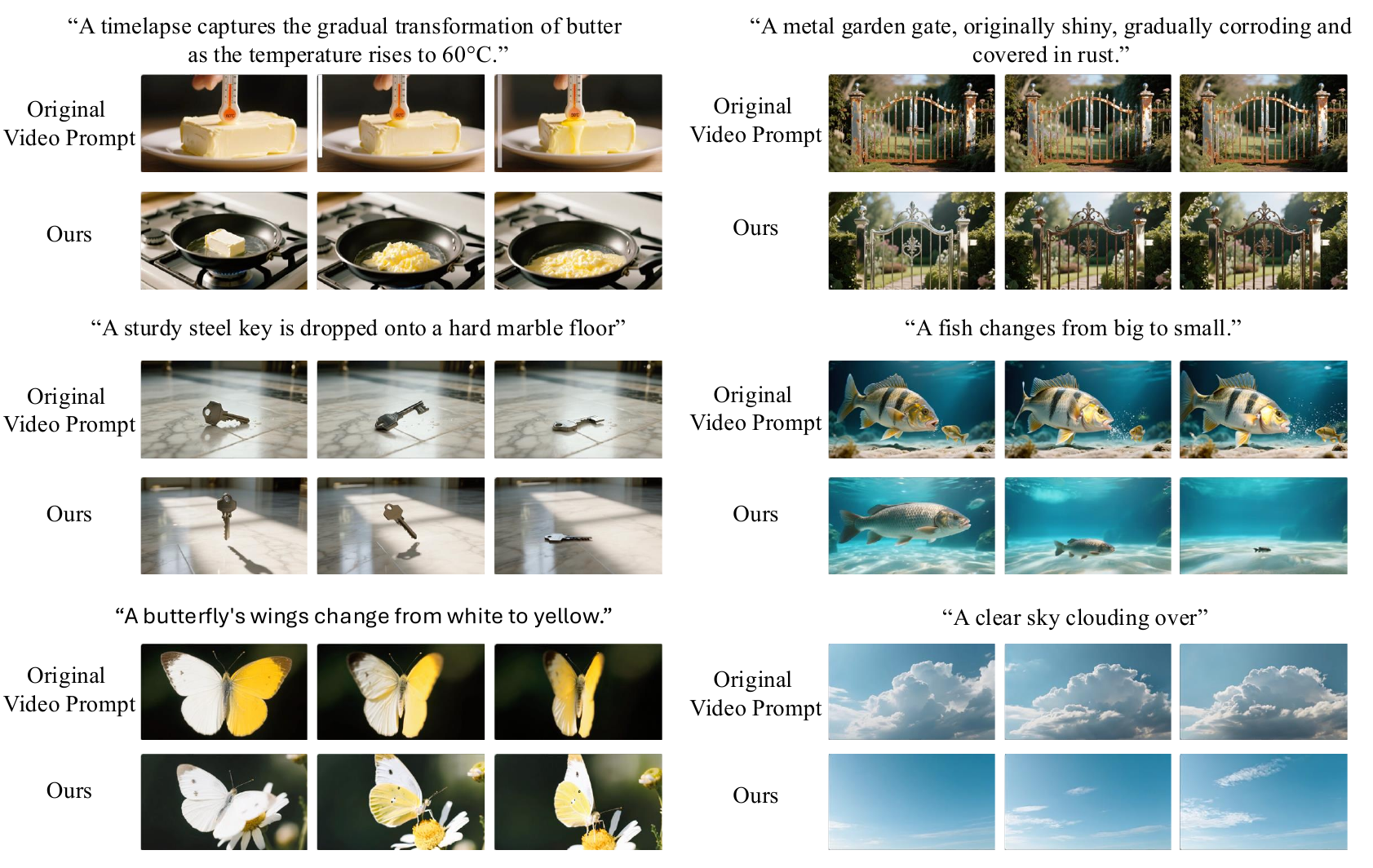}
\caption{Effect of LLM prompt rewriting. Anchor images generated from LLM-refined first-frame descriptions more accurately reflect the intended initial scene than anchors produced from raw prompts.}
\label{fig:llm}
\end{figure*}

\paragraph{Original Prompt vs.\ LLM-Refined First-Frame Prompt.}
Figure~\ref{fig:llm} compares anchor images generated directly from the raw video prompt with those produced using the LLM-refined first-frame description. The refined prompts generate anchors that more accurately capture the intended initial configuration of objects, spatial layout, and contextual cues. Videos produced from these anchors exhibit stronger compositional fidelity and temporal coherence. This qualitative trend aligns with the quantitative improvements reported in~\cref{ablation:caption}.

\begin{figure*}[t]
\label{fig:pull}
\centering
\includegraphics[width=\linewidth]{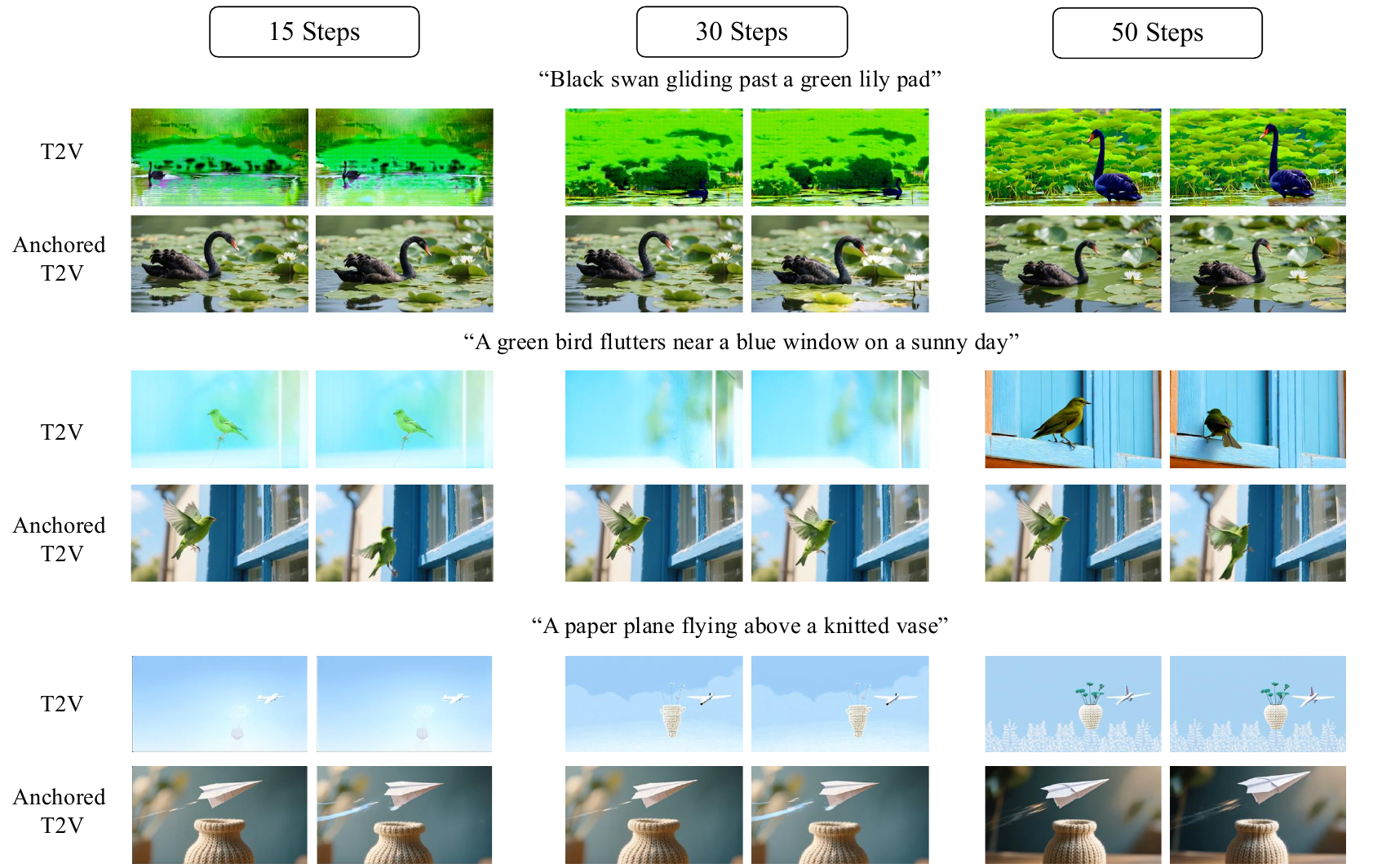}
\caption{Qualitative sensitivity to sampling steps. anchored T2V maintains coherent scene structure even at 15 steps, whereas text-only T2V collapses as sampling decreases.}
\label{fig:steps}
\end{figure*}

\paragraph{Effect of Reduced Sampling Steps.}
\Cref{fig:steps} illustrates how decreasing the number of diffusion steps affects visual quality. For text-only T2V models, fewer steps lead to degraded scene structure throughout the entire video. The anchored model preserves scene integrity even at 15 sampling steps. These results visually reinforce the stability trends observed quantitatively in~\cref{fig:qualitative}, demonstrating that anchoring makes denoising far more robust to aggressive sampling reduction.

\begin{figure*}[t]
\label{fig:pull}
\centering
\includegraphics[width=\linewidth]{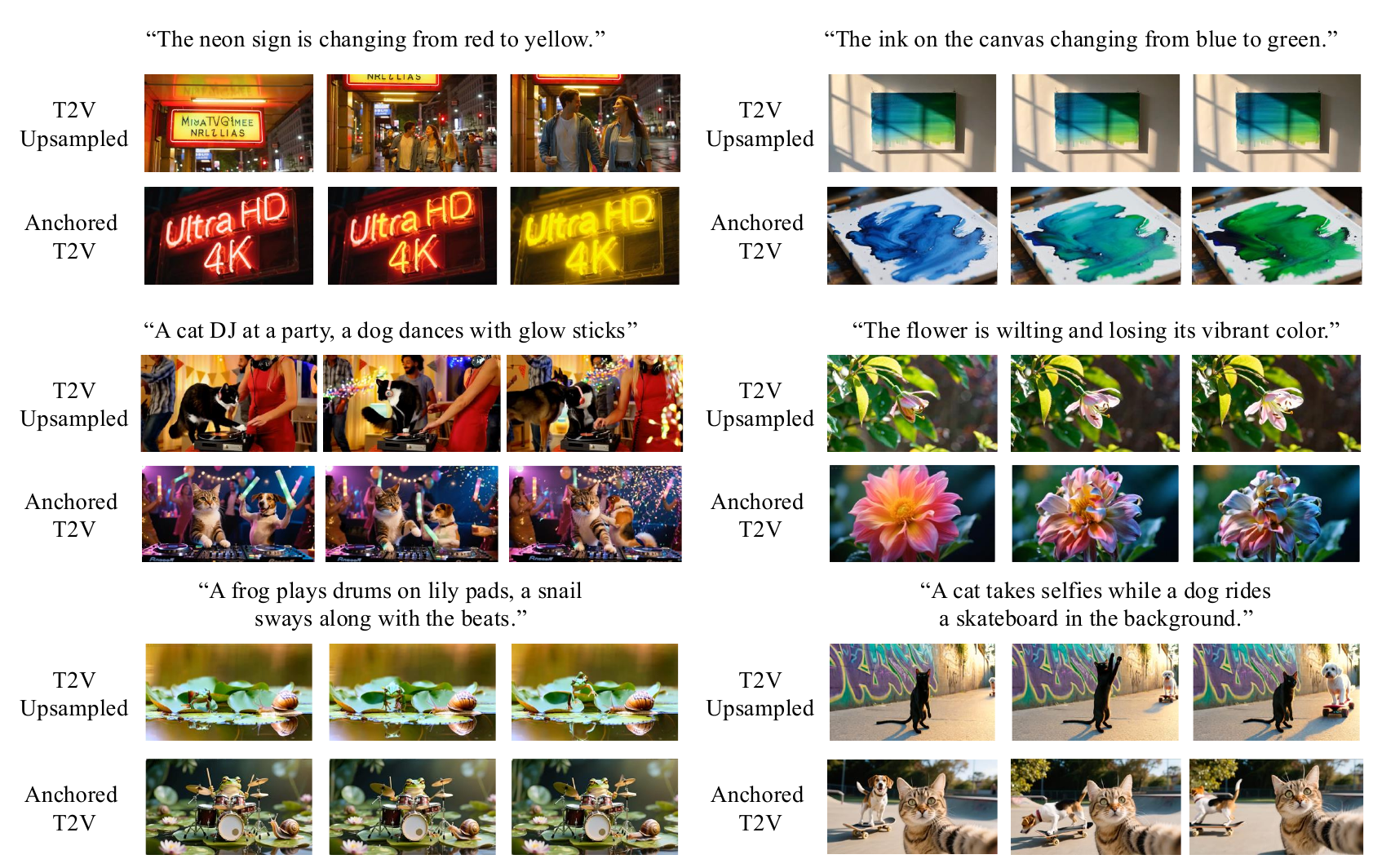}
\caption{Prompt upsampling vs.\ anchored T2V. Upsampled T2V improves textual detail but does not fix core compositional failures. Our anchored model successfully reconstructs the correct scene structure across the same prompts.}
\label{fig:upsampling}
\end{figure*}

\paragraph{Comparison with Upsampled Text}
Figure~\ref{fig:upsampling} illustrates typical failure modes of prompt upsampling. Although upsampling produces more detailed text descriptions, it does not correct the underlying compositional weaknesses of T2V models as they seem to lead to similar problems in terms of semantic composition. Our anchored model resolves these cases by grounding generation in a concrete visual anchor. These examples highlight that text refinement alone cannot substitute for explicit scene grounding.

% \section{Rationale}
% \label{sec:rationale}
% % 
% Having the supplementary compiled together with the main paper means that:
% % 
% \begin{itemize}
% \item The supplementary can back-reference sections of the main paper, for example, we can refer to \cref{sec:intro};
% \item The main paper can forward reference sub-sections within the supplementary explicitly (e.g. referring to a particular experiment); 
% \item When submitted to arXiv, the supplementary will already included at the end of the paper.
% \end{itemize}
% % 
% To split the supplementary pages from the main paper, you can use \href{https://support.apple.com/en-ca/guide/preview/prvw11793/mac#:~:text=Delete%20a%20page%20from%20a,or%20choose%20Edit%20%3E%20Delete).}{Preview (on macOS)}, \href{https://www.adobe.com/acrobat/how-to/delete-pages-from-pdf.html#:~:text=Choose%20%E2%80%9CTools%E2%80%9D%20%3E%20%E2%80%9COrganize,or%20pages%20from%20the%20file.}{Adobe Acrobat} (on all OSs), as well as \href{https://superuser.com/questions/517986/is-it-possible-to-delete-some-pages-of-a-pdf-document}{command line tools}.